\documentclass[lettersize,journal]{IEEEtran}
\usepackage{amsmath,amsfonts}
\usepackage{algorithm}
\usepackage{array}
\usepackage[caption=false,font=normalsize,labelfont=sf,textfont=sf]{subfig}
\usepackage{textcomp}
\usepackage{stfloats}
\usepackage{url}
\usepackage{verbatim}
\usepackage{graphicx}
\usepackage{tabularx}
\usepackage{xcolor,colortbl}
\usepackage{cite}
\usepackage{float}
\usepackage{algpseudocode}
\usepackage{multirow}

\usepackage{hhline} 
\usepackage{amssymb} 
\usepackage{pifont}
\usepackage{makecell}
\usepackage{cleveref}

\usepackage{arydshln}
\definecolor{lightgrey}{rgb}{0.85, 0.86, 0.85}
\definecolor{lightpurple}{rgb}{0.86, 0.87, 0.86}
\definecolor{fst}{rgb}{0.753, 0.886, 0.792}
\definecolor{sec}{rgb}{0.886, 0.929, 0.725}
\definecolor{thd}{rgb}{1.0, 0.980, 0.757}

\usepackage{utfsym}
\usepackage{xcolor,pifont}
\newcommand*\colourcheck[1]{%
  \expandafter\newcommand\csname #1check\endcsname{\textcolor{#1}{\ding{52}}}%
}
\colourcheck{green}

\begin{document}

\title{MG-SLAM: Structure Gaussian Splatting SLAM with Manhattan World Hypothesis}

\author{Shuhong Liu$^*$, Tianchen Deng$^*$\thanks{$^*$ These authors contributed equally to this work.}, Heng Zhou, Liuzhuozheng Li, Hongyu Wang \textit{Member, IEEE},\\
Danwei Wang \textit{Fellow, IEEE}, Mingrui Li$^{\dag}$ \thanks{$^{\dag}$ Corresponding author: mmclmr@mail.dlut.edu.cn}
\thanks{Shuhong Liu and  Liuzhuozheng Li are with Department of Information Science and Technology and Department of Complexity Science and Engineering, The University of Tokyo, Tokyo 113-8654, Japan. Tianchen Deng is with Institute of Medical Robotics and Department of Automation, Shanghai Jiao Tong University, and Key Laboratory of System Control and Information Processing, Ministry of Education, Shanghai 200240, China. Heng Zhou is with Department of Mechanical Engineering, Columbia University, New York 10027, United States. Danwei Wang is with School of Electrical and Electronic Engineering, Nanyang Technological University, Singapore. Hongyu Wang and Mingrui Li are with Department of Computer Science, Dalian University of Technology, Dalian 116024, China. This research is supported by the National Research Foundation, Singapore, under the NRF Medium Sized Centre scheme (CARTIN), ASTAR under National Robotics Programme with Grant No. M22NBK0109, and by the National Research Foundation, Singapore. Any opinions, findings and conclusions or recommendations expressed in this material are those of the authors and do not reflect the views of National Research Foundation.}}

\markboth{IEEE Transactions on Automation Science and Engineering}%
{Shell \MakeLowercase{\textit{et al.}}: A Sample Article Using IEEEtran.cls for IEEE Journals}


\maketitle

\begin{abstract}
Gaussian Splatting SLAMs have made significant advancements in improving the efficiency and fidelity of real-time reconstructions. However, these systems often encounter incomplete reconstructions in complex indoor environments, characterized by substantial holes due to unobserved geometry caused by obstacles or limited view angles. To address this challenge, we present Manhattan Gaussian SLAM, an RGB-D system that leverages the Manhattan World hypothesis to enhance geometric accuracy and completeness. By seamlessly integrating fused line segments derived from structured scenes, our method ensures robust tracking in textureless indoor areas. Moreover, The extracted lines and planar surface assumption allow strategic interpolation of new Gaussians in regions of missing geometry, enabling efficient scene completion. Extensive experiments conducted on both synthetic and real-world scenes demonstrate that these advancements enable our method to achieve state-of-the-art performance, marking a substantial improvement in the capabilities of Gaussian SLAM systems. \\

\textit{Note to Practitioners}---This paper was motivated by the limitations of Gaussian Splatting SLAM systems in complex indoor environments, where textureless surfaces and obstructed views often lead to substantial tracking errors and incomplete maps. While existing systems excel in high-fidelity reconstruction, they struggle with frame-to-frame or point-feature tracking in large-scale environments, particularly with significant camera rotations and obscured structures. In this paper, we enhance the neural dense SLAM by integrating the Manhattan World hypothesis, applying its parallel line and planar surface constraints for more robust tracking and mapping. We incorporate line segment features into both tracking and mapping to improve structural accuracy. Moreover, we propose a post-optimization method that interpolates new Gaussian primitives to effectively fill gaps on planar surfaces. Extensive experiments on multiple datasets demonstrate the superiority of our approach in large-scale indoor environments,  resulting in more accurate tracking and mapping.
\end{abstract}


\section{Introduction}

Simultaneous Localization and Mapping (SLAM) is a fundamental problem in computer vision that aims to map an environment while simultaneously tracking the camera pose. Learning-based dense SLAM methods, particularly neural radiance field (NeRF) approaches \cite{zhu2022nice, kong2023vmap, wang2023co, johari2023eslam, zhu2023nicer, sandstrom2023point, deng2024plgslam, deng2024neslam, liu2025i2nerf}, have demonstrated remarkable improvements in capturing dense photometric information and providing accurate global reconstruction over traditional systems based on sparse point clouds \cite{newcombe2011dtam, salas2013slam++, mur2015orb, gao2019autonomous, freda2023plvs, wang2024lfvislam, lin2024robust}. However, NeRF methods still face drawbacks such as over-smoothing, bounded scene representation, and computational inefficiencies \cite{xu2024gridguided, yan2024mvoxti, li2024dgnr, zhang2025efficient}. Recently, Gaussian-based SLAM \cite{keetha2023splatam, matsuki2023gaussian, yugay2023gaussian, li2024sgs, huang2024photo, deng2024compact, lidense2025} has emerged as a promising approach utilizing volumetric Gaussian primitives \cite{kerbl20233d}. Leveraging these explicit representations, Gaussian SLAMs deliver high-fidelity rendering and fine-grained scene reconstruction, overcoming the limitations of NeRF-based methods.

\begin{figure}[t]
    \begin{center}
        \includegraphics[width=0.40\textwidth]{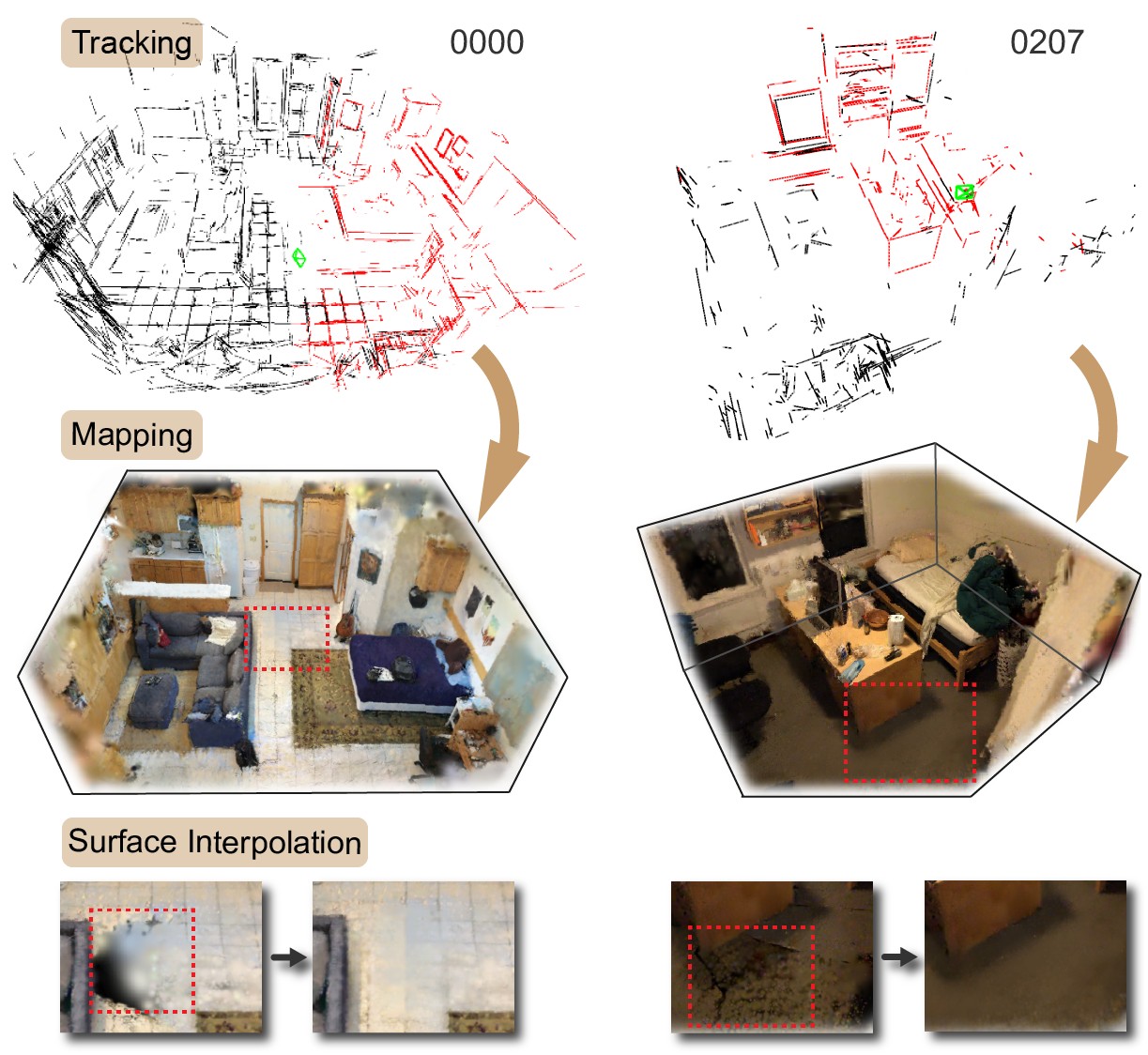}
    \end{center}
    \caption{Visualization of MG-SLAM on \textit{scene0000\_00} and \textit{scene0207\_00} of the ScanNet dataset \cite{dai2017scannet}. Our method leverages robust line segments to achieve superior camera pose estimation and scene reconstruction results. Moreover, by applying structural surface constraints, we enhance and complete the planar surfaces of the scene through the insertion of new Gaussian primitives to fill gaps.}
    \label{fig:demo}
\end{figure}

Despite their strengths, Gaussian SLAM faces notable challenges in indoor scenes, which are often characterized by textureless surfaces and complex spatial layouts. These environments hinder robust tracking due to a lack of sufficient texture details critical for camera pose optimization. Moreover, the complex geometry of indoor scenes often leads to substantial unobserved areas due to occlusions or limited view coverage. These unseen regions pose a critical yet largely unexplored challenge for Gaussian SLAM, as the Gaussian representation can hardly interpolate the unobserved geometry without multi-view optimization. Consequently, substantial holes and gaps are left in the unseen areas of the map, an issue that has been largely overlooked in previous Gaussian SLAM studies.

To overcome these challenges, we leverage the renowned Manhattan World (MW) hypothesis \cite{coughlan2000manhattan} as a foundational strategy for refining and completing scene geometries. This assumption posits that the built environment predominantly adheres to a grid-like structure, with surfaces and lines aligning with three orthogonal directions. These lines and planar surfaces impose meaningful constraints on the tracking and mapping processes in the Gaussian SLAM system.

Specifically, we encompass enhancements in tracking, mapping, and planar surface completion. In tracking, we utilize fused line features derived from the structured scenes as robust feature foundations in textureless areas, backprojecting and reprojecting these line segments for pose optimization and full bundle adjustment. In mapping, we apply a photometric loss for the reprojected line features to refine the map. This approach ensures that the reconstructed scene adheres closely to the true structure of the environment, thereby improving both its geometry accuracy and rendering quality.

Furthermore, the MW hypothesis facilitates the identification and interpolation of structured surfaces, such as floors and ceilings. These planar surfaces are critical to defining the overall geometry of indoor scenes but are often partially obscured or missing from the captured views \cite{guo2022neural,zhou2024neural}. By segmenting these incomplete surfaces---refined by the extracted lines as boundaries---we can predict their continuation beyond the directly observed portions by generating new Gaussians. This strategy enables us to optimize the representation of large surfaces within the scene, enhancing the completeness of the rendered map. Finally, we compress the Gaussian representation into mesh surfaces by incorporating regularization terms through Poisson reconstruction \cite{guedon2023sugar}. This approach enables the extraction of high-quality mesh, previously unavailable in the Gaussian SLAM systems, making it readily available for downstream tasks.

Overall, our work presents the following key contributions:
\begin{itemize}
\item We propose MG-SLAM, a novel RGB-D Gaussian SLAM system that capitalizes on the MW hypothesis. This assumption introduces lines and planar surfaces for robust tracking, map refinement, and surface completion for neural-dense SLAM systems.

\item We incorporate line segments along with an additional fusion and filtering strategy into the neural-dense SLAM system, effectively improving its tracking capabilities in textureless indoor environments and enhancing the quality of the dense Gaussian map.

\item We establish hypothesis surfaces using extracted line segments that represent planar boundaries. These surfaces guide our efficient interpolation of new Gaussians to fill gaps and holes in the reconstructed map, seamlessly addressing areas where current Gaussian SLAM systems face limitations due to unobserved geometry.

\item Extensive experiments conducted on both large-scale synthetic and real-world datasets demonstrate that our system offers state-of-the-art (SOTA) tracking and comprehensive map reconstruction, achieving 50\% lower ATE and 5dB enhancement in PSNR on large-scale Apartment dataset, meanwhile operating at a high frame rate.
\end{itemize}

\section{Related Work}

\subsection{Neural Dense SLAM}
Neural Implicit SLAM systems \cite{zhu2022nice, li2023end, johari2023eslam, wang2023co, sandstrom2023point, zhang2023go, zhang2023hi, liso2024loopy, zhou2024mod} leveraging NeRF \cite{mccormac2018fusion} are adept at handling complex reconstruction using implicit volumetric representation. Despite these advancements, NeRF methods often struggle with the over-smoothing issue, where fine-grained object-level geometry and features are difficult to capture during reconstruction \cite{li2024sgs}. Moreover, these methods suffer from catastrophic loss as the scenes are implicitly represented by MLPs.

In contrast, the high-fidelity and fast rasterization capabilities of 3D Gaussian Splatting \cite{kerbl20233d} enable higher quality and efficiency on scene reconstruction \cite{keetha2023splatam, yan2023gs, matsuki2023gaussian, huang2024photo, peng2024rtg, li2024sgs, tosi2024nerfs, li2024ngm, zhu2024loopsplat, liu2025deraings, liu2025realx3d, ha2025rgbd, hu2025cg}. MonoGS \cite{matsuki2023gaussian} utilizes a map-centric SLAM approach that employs 3D Gaussian representation for dynamic and high-fidelity scene capture. SplaTAM \cite{keetha2023splatam} adopts an explicit volumetric approach using isotropic Gaussians, enabling precise camera tracking and map densification. Photo-SLAM \cite{huang2024photo} and RTG-SLAM \cite{peng2024rtg} integrate the traditional feature-based tracking system \cite{mur2017orb, chung2023orbeez} with Gaussian mapping, providing robust tracking and exceptional real-time processing capabilities. However, existing Gaussian SLAM systems lack effective camera pose optimization in textureless environments, limiting tracking accuracy in indoor scenes. Moreover, they struggle to effectively reconstruct unobserved areas, often resulting in incomplete reconstructions with gaps and holes. This limitation becomes more problematic in settings where the camera’s movement is restricted, leading to significant unmodeled areas in structured indoor scenes. Additionally, current Gaussian-based SLAM approaches face challenges in direct mesh generation due to the discrete nature of 3D Gaussian primitives, which complicates surface extraction. To address this, recent Gaussian-based systems \cite{yan2023gs, yugay2023gaussian, zhu2024loopsplat} apply TSDF fusion \cite{curless1996volumetric} on rendered images to produce meshes. However, this approach is limited to observed viewpoints, leaving unobserved regions incomplete. More critically, it fundamentally depends on the offline volumetric TSDF projection using rendered images, which is independent of the reconstructed Gaussian maps. To overcome these challenges, our method incorporates line segments and planar surface assumption to seamlessly fill the gaps of structured surfaces and directly extract high-quality mesh from volumetric Gaussian representations. 

\subsection{SLAM with Structure Optimization}
Line features are known to significantly enhance camera pose optimization by capturing high-level geometric elements and structural properties \cite{zhou2022edplvo, xu2023plpl}. Traditional SLAM systems \cite{sun2021plane, xu2022eplf, freda2023plvs, shu2023structure, zhao2023visual, chen2024vpl, jiang2024ulslam, wang2025mssd} combine point features, line segments, or planar surfaces to refine camera pose estimation and improve map reconstruction. These approaches are particularly effective in dynamic or textureless environments, where point-based methods often face substantial challenges. 

Building on the MW hypothesis, recent advancements in SLAM systems leveraging planar constraints have further mitigated tracking and mapping drift. For instance, \cite{li2018monocular} demonstrates accurate camera pose estimation and sparse 3D map generation in monocular SLAM. \cite{liu2020visual} achieves drift-free rotational motion estimation by leveraging structural regularities captured by line features. ManhattanSLAM \cite{yunus2021manhattanslam} extends this concept by optimizing camera trajectories and generating sparse maps with points, lines, and planes, alongside dense surfel-based reconstructions. Similarly, \cite{peng2021accurate} incorporates Manhattan frame re-identification to build robust rotational constraints, which are tightly integrated into a bundle adjustment framework. Planar-SLAM \cite{li2021rgb} focuses on planar mesh reconstruction by utilizing line and sparse point features, while \cite{li2022dr} introduces drift-free rotation estimation through the use of Gaussian spheres. Moreover, \cite{zhang2023visual} refines spatial constraints through advanced bundle adjustment leveraging structure constraints. \cite{jeong2023linear} integrates low-cost LiDAR in structured scenes for better view coverage.

Despite these advancements, most existing systems leveraging structure or planar constraints rely on sparse map representations, such as point clouds or simple planar surfels, which are insufficient for reconstructing fine-grained maps with detailed textures. Additionally, the sparse map leads to discontinuities in the reconstructed map, making surface extraction and subsequent optimization difficult. To bridge this gap, we incorporate line features into the neural dense system, enhancing its tracking capabilities in indoor environments and enabling structured surface optimization.

\begin{figure*}[!tp]
    \begin{center}
        \includegraphics[width=\textwidth]{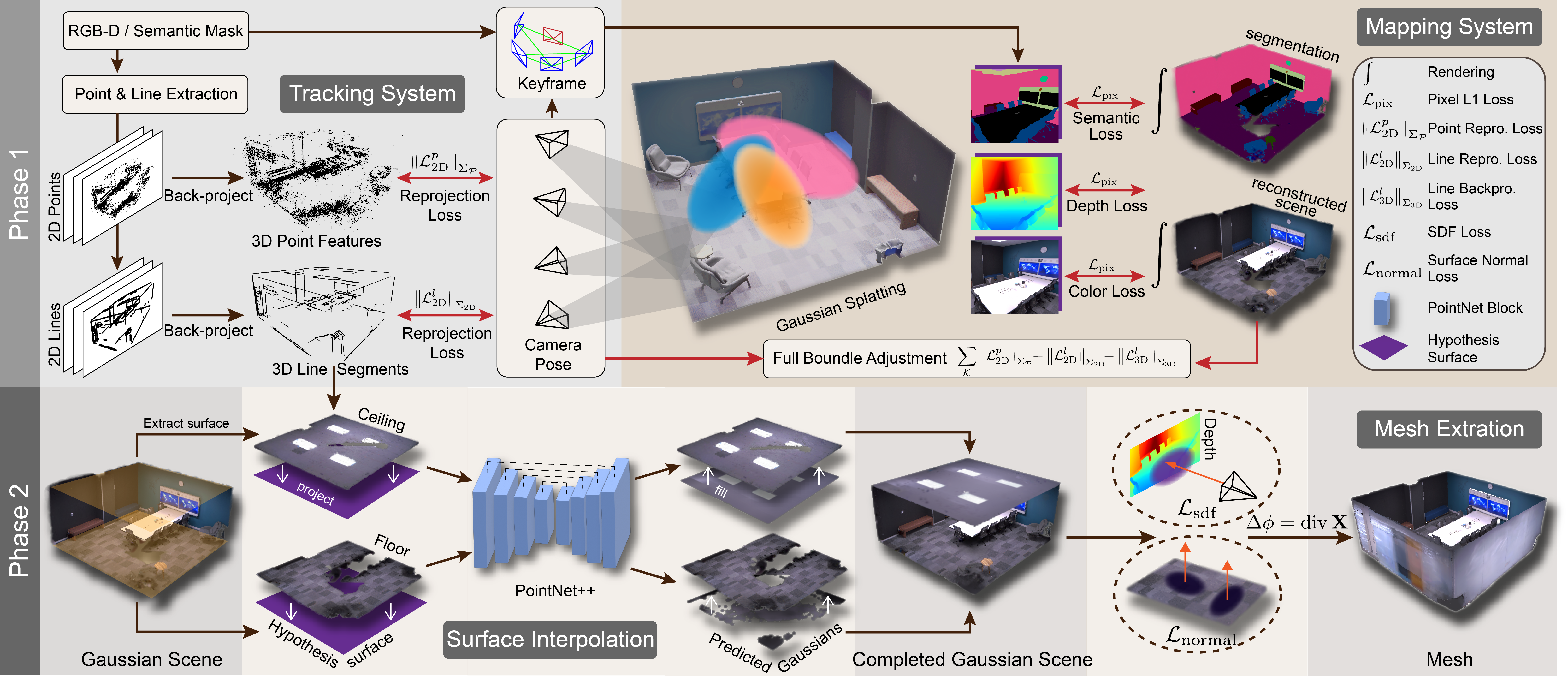}
    \end{center}
    \caption{The two-phase pipeline illustration of our proposed MG-SLAM. The upper section visualizes the parallel processes of the tracking and mapping systems. The lower section presents the post-optimization of scene interpolation and mesh extraction.}
    \label{fig:pipeline}
\end{figure*}

\section{Method}
\Cref{fig:pipeline} illustrates the pipeline of our proposed method. Under the constraints of the Manhattan World, MG-SLAM introduces line segments and structured surfaces to enhance camera pose estimation and map reconstruction. \Cref{sec:tracking} details the tracking mechanism that utilizes both point and line features. We utilize a specific strategy for fusing line segments to ensure reliable identification of line features. \Cref{sec:mapping} discusses the Gaussian representation, including a specialized loss term dedicated to the reconstruction of line segments. \Cref{sec:structure_opt} describes the completion and refinement of the scene, grounded in the assumption of structured surfaces. \Cref{sec:mesh_generate} describes the mesh generation utilizing regularization losses.

\subsection{Notation}
We define the 2D image domain as $\mathcal{P} \in \mathbb{R}^2$, which encompasses appearance color information $\mathcal{C} \in \mathbb{N}^3$, semantic color $\mathcal{S} \in \mathbb{N}^3$, and depth data $\mathcal{D} \in \mathbb{R}^{+}$. Transitioning to the 3D world domain, denoted as $\mathcal{X} \in \mathbb{R}^3$, we introduce a camera projection function $\pi: \mathcal{X} \rightarrow \mathcal{P}$, mapping a 3D point $\mathcal{X}_i$ to its 2D counterpart $\mathcal{P}_i$, and conversely, a backprojection function $\theta: \mathcal{P} \rightarrow \mathcal{X}$, for the reverse mapping. $(R, t) \in SE\left( 3 \right)$ defines the camera pose. $\mathbb{I}$ is the identity matrix.

\subsection{Tracking}
\label{sec:tracking}

We utilize the backprojection of point features and line segments extracted from 2D images into 3D space in parallel for optimizing the camera pose, based on PLVS \cite{freda2023plvs}. Moreover, we propose strategies for line segment fusion and suppression, designed to merge shorter segments into longer ones and eliminate unstable lines. This method is grounded in the understanding that longer line features tend to offer greater reliability and robustness throughout the tracking process.

\subsubsection{Point Reprojection Error}
Given a point observation $p_i$ within a specific keyframe $k$, its reprojection error can be computed as:
\begin{equation}
\mathcal{L}_{\rm 2D}^{p} = \left\Vert \mathcal{P}_i - \pi(R_k\mathcal{X}^w_i+ T_k) \right\Vert_{\Sigma_{\mathcal{P}_i}} \in \mathbb{R}
\end{equation}

\noindent where $\mathcal{X}^w_i$ is the 3D point back-projected to the world domain and transformed to the camera frame $\mathcal{X}^c_i$. The term $\Sigma_{\mathcal{P}i}=\sigma^2{\mathcal{P}_i}\mathbb{I}_2$ represents the covariance matrix, encapsulating the variance $\sigma^2{\mathcal{P}_i}$ of point detection noise at different scales within a Gaussian image pyramid \cite{freda2023plvs}.

\subsubsection{Line Reprojection and Backprojection Error}
For line segments, represented in the 3D world as a pair of endpoints $(\mathcal{X}_p, \mathcal{X}_q) \in \mathbb{R}^3$ and in the image plane as $(\mathcal{P}_p, \mathcal{P}_q) \in \mathbb{R}^2$, we identify the line segments on each level of the image pyramid using the EDlines method \cite{akinlar2011edlines} for efficient extraction. Matching line segments leverage the Line Band Descriptor (LBD) method \cite{zhang2013efficient}, and matches are scored according to the Hamming distances between descriptors.

Upon observing a line segment $l = (\mathcal{P}_p, \mathcal{P}_q)_k$, its back-projected 3D endpoints can be computed as $L = (\theta(\mathcal{P}_p), \theta(\mathcal{P}_q))_k$. Then the reprojection error of the line segment can be interpreted as:
\begin{align}
\mathcal{L}_{\rm 2D}^{l} &= \begin{bmatrix}
\left\| d_k \cdot \pi(R_k \mathcal{X}_p^w + t) \right\| \\
\left\| d_k \cdot \pi(R_k \mathcal{X}_q^w + t) \right\|
\end{bmatrix}_{\Sigma_{\rm 2D}} \in \mathbb{R}^2~,\\
d_k &= \frac{\bar{\mathcal{P}_p} \times \bar{\mathcal{P}_q}}{\left\| \bar{\mathcal{P}_p} \times \bar{\mathcal{P}_q} \right\|} |_k
\end{align}

\noindent where $\bar{\mathcal{P}_p} = [\mathcal{P}_p^T, 1]^T$, $\bar{\mathcal{P}_q} = [\mathcal{P}_q^T, 1]^T$. The covariance matrix $\Sigma_{\rm 2D} = \sigma^2_{d_i}\mathbb{I}_2$ is assumed to be diagonal for simplicity. Further, we calculate the perpendicular distance in 3D between a map point $\mathcal{X}^w_i \in \mathbb{R}^3$ and a 3D line segment $L$ as follows:
\begin{equation}
e_{\rm 3D}^l(L, \mathcal{X}_i^c)_k = \frac{\left\Vert (\mathcal{X}^c_i - \theta(\mathcal{P}_p)) \times (\mathcal{X}^c_i - \theta(\mathcal{P}_q)) \right\Vert_k}{\left\Vert \theta(\mathcal{P}_p) - \theta(\mathcal{P}_q) \right\Vert_k}
\label{eq:line_distance}
\end{equation}

\noindent which considers the cross-product of the differences between the point's position and the back-projected positions of the line's endpoints, normalized by the distance between these endpoints. Additionally, the distance between the world domain point $\mathcal{X}^w_i$ and its corresponding back-projected point in the image plane $\mathcal{P}_i \in \mathbb{R}^2$ is determined by:
\begin{equation}
e_\mathcal{P}^l(\mathcal{P}_i, \mathcal{X}_i^c) = \left\Vert \mathcal{X}^c_i - \theta(\mathcal{P}_i) \right\Vert_k~,~~\mathcal{X}^c_i = R_k\mathcal{X}^w_i + T_k
\label{eq:point_distance}
\end{equation}

\noindent Consequently, the back projection error is formulated as:
\begin{equation}
\mathcal{L}_{\rm 3D}^{l} =
\begin{bmatrix}
e_{\rm 3D}(\theta(\mathcal{P}_i), \mathcal{X}^{w}_{p}) + \beta \cdot e_{\mathcal{P}}(\mathcal{P}_i, \mathcal{X}^{w}_{p}) \\
e_{\rm 3D}(\theta(\mathcal{P}_i), \mathcal{X}^{w}_{q}) + \beta \cdot e_{\mathcal{P}}(\mathcal{P}_i, \mathcal{X}^{w}_{q})
\end{bmatrix}_k \in \mathbb{R}^2
\end{equation}

\noindent In this context, $\beta \in [0, 1]$ acts as a weighting parameter, ensuring the endpoints of the 3D line segment remain stable throughout the optimization process, thereby preventing drift.

\subsubsection{Line Segment Fusion}

The stability of tracking can be notably enhanced by the presence of elongated line segments. However, the extraction process, particularly through EDLines, often yields fragmented segments as a consequence of various disturbances like image noise. Our method integrates the following key steps: (1) Fusing line segments that are directionally aligned within a one-degree angle difference, ensuring they follow the same path. (2) Ensuring these segments are close—within a 10-pixel distance of each other's nearest endpoints—yet do not overlap, preserving their distinctness while allowing for precise merging. (3) Verifying the vertical distance between the corresponding endpoints of one segment to the entirety of another is less than certain pixel threshold, maintaining geometric consistency. Only segments fulfilling all three criteria are merged, producing longer and more reliable lines. Furthermore, we filter out segments that fall below a predefined length threshold, relative to the image size, to improve the system's tracking robustness.

\subsubsection{Full Bundle Adjustment Error}
The overall objective function for the full bundle adjustment is given by:
\begin{align}
\mathcal{L}_{\rm BA} = \sum_{\mathcal{K}} \sum_{\mathcal{U}} \rho \left( \left\| \mathcal{L}_{\rm 2D}^{p} \right\|_{\Sigma_{\mathcal{P}}} \right) &+
\sum_{\mathcal{V}} \rho \left( \left\| \mathcal{L}_{\rm 2D}^{l} \right\|_{\Sigma_{\rm 2D}} \right) \nonumber \\
&+ \sum_{\mathcal{V}} \rho \left( \left\| \mathcal{L}_{\rm 3D}^{l} \right\|_{\Sigma_{\rm 3D}} \right)
\label{eq:ba_error}  
\end{align}

\noindent Here, $\mathcal{K}$ represents the set of chosen keyframes, while $\mathcal{U}$ and $\mathcal{V}$ denote the sets of points and lines extracted within the current frame. The Huber cost function $\rho(\left\Vert e \right\Vert)_{\Sigma^{-1}}) = e^T\Sigma^{-1}e$ is applied to mitigate the influence of outliers. The optimization process utilizes the Levenberg-Marquardt method \cite{levenberg1944method, marquardt1963algorithm}, which solves the augmented normal equations by iteratively updating the parameters $\Theta$ as:
\begin{equation}
(A_{\epsilon}^T \Sigma_{\epsilon}^{-1} A_{\epsilon} + \Lambda \mathbb{I}) \Delta \Theta = -A_{\epsilon}^T \Sigma_{\epsilon}^{-1} \epsilon
\end{equation}

\subsection{Mapping}
\label{sec:mapping}

The map representation is based on 3D Gaussians primitives $\mathcal{G} = \alpha \mathcal{N}(\mu^w, \Sigma^w)$, where $\alpha \in [0, 1]$ is the opacity, $\mu^w$ and $\Sigma^w$ are mean and covariance in world coordinate. Each $\mathcal{G}_i$ is associated with the color of appearance and semantic feature $f_i = \{c_i \in \mathbb{R}^3, s_i\in \mathbb{R}^3\}$. Semantic segmentation is utilized to identify the surface such as floors for structure optimization and surface extraction, explained in \Cref{sec:structure_opt}.

\subsubsection{Scene Representation}
We use the standard point rendering formula \cite{yifan2019differentiable,kopanas2021point} to splat $\mathcal{G}$ to render 2D image:
\begin{equation}
\mu_{\mathcal{P}} = \pi(R\mu_{\mathcal{G}} + T),~\Sigma_{\mathcal{P}} = J R \Sigma_{\mathcal{G}} R^T J^T
\label{eq:scene_representation}
\end{equation}

\noindent where $\mu_{\mathcal{G}}$ and $\Sigma_{\mathcal{G}}$ are mean and covariance of the Gaussian primitives. $J$ is the Jacobian of the linear approximation of $\pi(\cdot)$. For each pixel ${\rm pix} \in \mathcal{P}$, the influence of $N$ Gaussians on this pixel can be combined by sorting the Gaussians in depth order and performing front-to-back alpha-blending:
\begin{equation}
\mathcal{P}_{\rm pix} = \sum_{i \in N} f_i \alpha_i \prod_{j=1}^{i-1} (1 - \alpha_j),
~\mathcal{D}_{\rm pix} = \sum_{i \in N} z_i \alpha_i \prod_{j=1}^{i-1} (1 - \alpha_j)
\end{equation}

\noindent where $\mathcal{P}_{\rm pix}$ and $\mathcal{D}_{\rm pix}$ represent the pixel-wise appearance ($\mathcal{C}_{\rm pix}$ for color and $\mathcal{S}_{\rm pix}$ for semantic features) and depth respectively. $z$ is the distance to the mean of the Gaussian $\mathcal{G}$ along the camera ray. 

\subsubsection{Mapping Loss}
\label{sec:mapping_loss}

The scene representation is optimized using keyframes obtained from the tracking system by minimizing the photometric residual:
\begin{align}
\mathcal{L}_{\rm pix} =  \left\Vert \mathcal{C}_{\rm pix} - \mathcal{C}^{\rm GT}_{\rm pix} \right\Vert &+ \lambda_{\mathcal{S}} \left\Vert \mathcal{S}_{\rm pix} - \mathcal{S}^{\rm GT}_{\rm pix} \right\Vert \nonumber \\
&+ \lambda_{\mathcal{D}} \left\Vert \mathcal{D}_{\rm pix} - \mathcal{D}^{\rm GT}_{\rm pix} \right\Vert
\label{eq:pixel_loss} 
\end{align}

\noindent In this context, Gaussian primitives are optimized to adjust their optical and geometrical parameters to closely replicate the observed scene with fine-grained details. Furthermore, to impose constraints on the scene utilizing the identified line features that precisely delineate the lines and potential edges in the current view, an additional line residual is introduced for pixels associated with line features to enhance the map's accuracy. Consequently, the aggregate loss for mapping is determined as follows:
\begin{equation}
\mathcal{L}_{\rm Mapping} = \sum_{\rm pix \in \mathcal{P}} \mathcal{L}_{\rm pix} + \lambda_\ell \sum_{{\rm pix} \in \mathcal{V}} \mathcal{L}_{\rm pix}
\label{eq:map_loss}
\end{equation}

$\lambda_{\mathcal{S}}$, $\lambda_{\mathcal{D}}$ and $\lambda_\ell$ in \Cref{eq:pixel_loss} and \Cref{eq:map_loss} are weighting parameters.

\subsection{Structure Optimization}
\label{sec:structure_opt}

In our optimization process, we focus particularly on refining the representation of expansive surfaces that formulate the indoor scene, such as the floor and ceiling. By applying the MW assumption, we introduce the planar hypothesis surfaces that are informed by the structural regularity of the extracted 3D line features. These constraints are employed to identify surfaces that are not adequately represented and to interpolate new Gaussians for densifying or filling in gaps with textures that are reasonably predicted.

\subsubsection{Gaussian Density}
In our Gaussian map, we define the density of the Gaussian $\nu:\mathbb{R}^{3}\rightarrow\mathbb{R}_{+} $ as the sum of the Gaussian values weighted by their alpha-blending coefficients at any given grid points \textit{p} as: 
\begin{equation}
\nu(p) = \sum\limits_{\mathcal{G}_i}{\alpha_{\mathcal{G}_i}}\mathrm{exp}(-\frac{1}{2}(p-\mu_{\mathcal{G}_i}^{T})\Sigma_{\mathcal{G}_i}^{-1}(p-\mu_{\mathcal{G}_i})~,
\label{eq:sub_density}
\end{equation}

\noindent where the $\mu_{\mathcal{G}_i}, \Sigma_{\mathcal{G}_i},\alpha_{\mathcal{G}_i}$ are the means, covariances, and alpha-blending coefficients of the Gaussians $\mathcal{G}$. To simplify the calculation, $\nu(p)$ can be approximated following \cite{guedon2023sugar} as:
\begin{equation}
\nu^*(p) = \alpha_{\mathcal{G}^*} \mathrm{exp}(-\frac{1}{2}(p-\mu_{\mathcal{G}^*}^{T})\Sigma_{\mathcal{G}^*}^{-1}(p-\mu_{\mathcal{G}^*})~,
\label{eq:sub_aprrox_density}
\end{equation}

\noindent where $\mathcal{G}^*$ is the nearest Gaussian that contributes most to $p$. This Gaussian density function $\nu^*(\cdot)$ facilitates the subsequent identification of under-represented regions on the planar surfaces and the mesh generation process.

\subsubsection{Map Calibration}
Since the SLAM system relies on the initial camera pose as a reference frame, the reconstructed scenes usually do not satisfy the orthogonality assumption of the Manhattan World. To overcome this issue, we calibrate the reconstructed scene by applying a calibrating matrix $K$ to the coordinates and covariance of the Gaussians. This matrix $K$ is derived from clustering the directions of 3D line segments that are presumed to align with the scene axes.

After aligning the structured surface boundaries to orthogonal directions, they extend across the $xy \rightarrow \mathbb{R}^2$ plane. We use the calibrated line segments to outline the rectangular boundary of the hypothesized planes, capitalizing on the dense line features commonly found at the corners of scenes. \Cref{alg:calibration} presents the detailed implementation.

\begin{algorithm}
\footnotesize
\label{algo:algorithm1}
\caption{The pseudo-code for calibration and identification of surface boundary}
\begin{algorithmic}[1]
\Require Line segments $L \in \mathbb{R}^3$ and the reconstructed Gaussian map $\mathcal{G}$

\State $L_{\rm filtered} \gets$ select $|\ell| > \mathcal{T}_{\ell}$ \Comment{Filter out small line segments}
\For{each line segment $\ell$ in $L_{\rm filtered}$}
    \State $\{direction\} \gets (\ell.end - \ell.start).normalized()$
    \Statex \Comment{Add direction vector}
\EndFor

\State $\Vec{x}, \Vec{y}, \Vec{z} \gets k\_means(\{direction\}, k=3)$  \Comment{Find axes directions}

\State $K \gets \mathbb{I} \in \mathbb{R}^4$ \Comment{Define calibration matrix}
\State $K[0:3, 0] \gets \Vec{x}$
\State $K[0:3, 1] \gets \Vec{y}$
\State $K[0:3, 2] \gets \Vec{z}$

\For{each $\mathcal{G}_i$ in $\mathcal{G}$}
    \State $\mu_{\mathcal{G}_i} \gets K \cdot \mu_{\mathcal{G}_i}$  \Comment{Calibrate coordinates of Gaussian}
    \State $\gamma_{\mathcal{G}_i} \gets K \cdot \gamma_{\mathcal{G}_i}$ \Comment{Calibrate rotation of Gaussian}
\EndFor
\For{each line segment $\ell$ in $L$}
    \State $\ell \gets K \cdot\ell$ \Comment{Calibrate line segment}
\EndFor

\State $\mathcal{B} \gets$ select maximum and minimum $x$ and $y$ for endpoints in $L$
\Statex \Comment{Define surface boundary}
\State \textbf{Output:} boundary of surface $\mathcal{B} \in \mathbb{R}^2$
\end{algorithmic}
\label{alg:calibration}
\end{algorithm}

\begin{algorithm}
\footnotesize
\caption{The pseudo-code for scene completion}
\begin{algorithmic}[1]
\Require The calibrated Gaussian scene $\mathcal{G}$, the mask $\mathcal{M}$ extracted from semantic segmentation, and the boundary of surface $\mathcal{B}$, threshold of density $\mathcal{T}_{density}$.
\State $\mathcal{M}_{tg} \gets$ semantic masks that represent target surfaces of the scene, e.g. floor
\For{each structure surface $\mathcal{G}_i$ in $\mathcal{G}(\mathcal{M}_{tg})$}
\State $\mathcal{G}_i \gets DBSCAN.fit(\mathcal{G}_i)$
\Statex \Comment{Apply DBSCAN \cite{ester1996density} to eliminate outliers}
\State $\bar{z} \gets avg(z)$ for $z \in \mu_{\mathcal{G}_i}$
\State $\Pi \gets grid(\mathcal{B})$ \Comment{Create the 2D grid of hypothesis surface}
\State $\mathcal{D}_{i} \gets$ compute the Gaussian density for each point of the grid in $\Pi$.
\If{$\mathcal{D}_{i} \leq \mathcal{T}_{density}$}
\State $\mu_{\rm pred} \gets$ generates new Gaussians centers
\EndIf
\State $B^{*}_{\mathcal{G}_i} \gets$ ramdomly sample Gaussians from $\mathcal{G}$ to form training batch. 
\For{each batch $B_i$ in $B^{*}_{\mathcal{G}_i}$}
\State $\mathcal{F}(\mu_{\mathcal{G}_i} | \mathcal{G}_i \in B_i) \gets$ train PointNet++ \cite{qi2017pointnet++} model on $B_i$
\EndFor
\State $\mathcal{C}_{\rm pred} \gets \mathcal{F}(\mu_{pred})$ \Comment{Predict the color of new Gaussians}
\State $\mathcal{G}_{\rm pred} \gets$ formulate new Gaussians using $\{ \mu_{\rm pred} \}$ and $\{ \mathcal{C}_{\rm pred} \}$
\EndFor
\State $\mathcal{G} \gets \mathcal{G}_{\rm pred}$ \Comment{Update the overall scene}
\State \textbf{Output:} structure optimized $\mathcal{G}$ 
\end{algorithmic}
\label{algo:surface_interpolation}
\end{algorithm}

\subsubsection{Surface Interpolation}
\Cref{algo:surface_interpolation} shows our surface interpolation strategy. Specifically, we utilize the semantic information, incorporated in \Cref{eq:pixel_loss}, for identifying the planar surfaces such as floors and ceilings. The chosen target Gaussians are then projected from the 3D space onto the 2D hypothesis $xy$ plane. Subsequently, we apply a density threshold to the sampling density function, defined in \Cref{eq:sub_aprrox_density}, to detect potential holes or gaps on the surface. New Gaussian primitives are generated at the identified gaps that fall below a density threshold. Finally, we train a PointNet++ model \cite{qi2017pointnet++}, fitting it to the presented surface to learn the presented color patterns and interpolate the texture color of the new Gaussians based on their spatial correlation. We empirically found that the PointNet++ network, which incorporates spatial correlation, is sufficiently effective in predicting the color patterns of new primitives on the gaps of structured surfaces. Given that these planar surfaces typically exhibit less texture than specific objects, employing more sophisticated models does not provide substantial benefits but increases computational expenses.

Through this method, we aim to seamlessly combine the detailed representation provided by Gaussians with the need for explicit scene geometry that accounts for unseen views, thereby achieving a more comprehensive and accurate reconstruction of planar surfaces in indoor environments.

\subsection{Mesh Generation}
\label{sec:mesh_generate}

We employed the surface extraction method proposed by \cite{guedon2023sugar} with a novel normal regularization term during the scene refinement. To compress the Gaussians to align closely with the surface, \cite{guedon2023sugar} forces the Gaussians to be flat by setting one of their scaling factors nearly to zero. Consequently, $\nu^*(p)$ can be further simplified to $\hat{\nu}^*(p) = \alpha_{\mathcal{G}^*} \mathrm{exp}\left( -{2s^{-2}_{\mathcal{G}^*}} \langle p-\mu_{\mathcal{G}^*}, n_{\mathcal{G}^*} \rangle^2 \right)$  where $s_{\mathcal{G}}$  represents the smallest scaling factor and $n_{\mathcal{G}}$ its corresponding normal. Here, $\alpha_{\mathcal{G}^*} = 1$ is set to avoid semi-transparent Gaussians.

Subsequently, a regularization term is used to align the SDF function derived from the density function with its estimation from the scene as:
\begin{equation}
\mathcal{L}_{\rm sdf} = \frac{1}{|\mathcal{X}_p|} \sum_{x \in \mathcal{X}_p} |sdf(x) - \hat{sdf}(x)|
\end{equation}

\noindent Here, $\mathcal{X}_p$ is the sampled 3D points and $sdf(x) = \pm s_{\mathcal{G}^*} \sqrt{-2 \log(\hat{\nu}^*(x))}$ represents the ideal SDF. The estimated SDF $\hat{sdf}(x)$  is determined by subtracting the depth of $x$ from the corresponding rendered depth map. This regularization term encourages closer alignment of the estimated surface with the observed depth.

Moreover, to derive smooth surfaces, particularly in flat and textureless floor areas, we adopt the MW hypothesis, which assumes that planar floors are orthogonal to the vertical axis. We enforce this constraint by introducing a normal regularization term for the estimated SDF:
\begin{equation}
\mathcal{L}_{\text{normal}} = \frac{1}{|\mathcal{X}_{\text{floor}}|} \sum_{x_{\text{2D}} \in \mathcal{P}_{\text{floor}}} \left| 1 - \hat{n} \cdot \frac{\nabla f(x)}{\|\nabla f(x)\|} \right|
\end{equation}

\noindent In this context, the projected points $x_{\rm 2D}$, which pass through the camera ray and fall within the floor area, are encouraged to align their normal to the ideal vector $\hat{n}=\langle 0, 0, 1 \rangle$. The set $\mathcal{P}_{\text{floor}}$ consists of image pixels identified as the floor using semantic segmentation.



\section{Experiment}

\subsection{Experiment Settings}

\noindent \textbf{Implementation Details}~Our tracking system is implemented in C++, while the mapping system uses Python3 and CUDA C. The experiments were conducted using a single RTX A100-80GB GPU and 24-core AMD EPYC 7402 processor. We use Adam optimizer for Gaussian representation optimization and network training. The hyperparameters for 3D Gaussian splitting \cite{kerbl20233d} are the same as the original paper. \Cref{tab:hyperparameter} presents the values for hyperparameters in our system. For Gaussian insertion, in each keyframe, we sampled $\mu_{\mathcal{G}}$ in \Cref{eq:scene_representation} from the rendered depth map $\mathcal{D}_{\rm pix}$ following the distribution of $\mathcal{N}(\mathcal{D}_{\rm pix}, 0.2\sigma_{\mathcal{D}})$. For uninitialized areas, we initialize Gaussians by sampling from $\mathcal{N}(\bar{\mathcal{D}}_{\rm pix}, 0.5\sigma_{\mathcal{D}})$ where $\bar{\mathcal{D}}_{\rm pix}$ is the mean value. For pruning, we remove the Gaussians with opacity less than 0.6. Our system incorporates a segmentation loss introduced in \cite{li2024sgs}, and we leverage this semantic information to effectively extract structured surfaces, such as floors and ceilings, from the reconstructed scenes. We subsequently train a PointNet++ \cite{qi2017pointnet++} network for each structure surface and interpolate the missing geometry by predicting the color of inserted Gaussians. \\

\noindent \textbf{Datasets}~We evaluate our method using two datasets: Replica \cite{straub2019replica}, a synthetic dataset, and ScanNet \cite{dai2017scannet}, a challenging real-world dataset. The large-scale Replica Apartment dataset used in our experiments was released by Tandem \cite{koestler2022tandem}. For the Replica dataset \cite{straub2019replica}, the ground-truth camera pose and semantic maps are obtained through Habitat simulation \cite{savva2019habitat}. In the case of the ScanNet dataset \cite{dai2017scannet}, the ground-truth camera poses are derived using BundleFusion \cite{dai2017bundlefusion}. Moreover, we further validate our method on a long-trajectory dataset collected using our physical platform \cite{deng2025mcn}. \\

\noindent \textbf{Metrics}~To assess the quality of the reconstruction, we employ metrics such as PSNR, SSIM, and LPIPS. For evaluating camera pose, we use the average absolute trajectory error (ATE RMSE). The real-time processing capability, essential for SLAM systems, is measured in frames per second (FPS). Best results are shaded as \colorbox{fst}{\textbf{first}}, \colorbox{sec}{\textbf{second}}, and \colorbox{thd}{\textbf{third}}. \\

\noindent \textbf{Baselines}~We evaluate our tracking and mapping results against state-of-the-art methods, including NeRF-based approaches such as NICE-SLAM \cite{zhu2022nice}, Co-SLAM \cite{wang2023co}, ESLAM \cite{johari2023eslam}, and Point-SLAM \cite{sandstrom2023point}, as well as recent Gaussian-based methods including SplaTAM \cite{keetha2023splatam}, MonoGS \cite{matsuki2023gaussian}, Photo-SLAM \cite{huang2024photo}, and RTG-SLAM \cite{peng2024rtg}. The results for MonoGS \cite{matsuki2023gaussian} were obtained using its RGB-D mode.

\begin{table}[!tp]
\caption{MG-SLAM hyperparameters}
\begin{center}
    \begin{tabularx}{0.46\textwidth}{lll} 
    \hline
    Symbol & Explanation & Value \\ 
    \hline
    $N_p$ & number of point features in tracking & 4e3 \\
    $N_\ell$ & number of line segments in tracking & 200 \\
    $r_{d}$ & downsample ratio for Gaussian initialization & 8 \\
    $n_{map}$ & mapping iteration & 150 \\
    $SH_{degree}$ & degree of the spherical harmonics & 0 \\
    $l_{feat}$ & learning rate for SH features & 2.5e-3 \\
    $l_{opacity}$ & learning rate for opacity & 0.05 \\
    $l_{scale}$ & learning rate for scaling & 1e-3 \\
    $l_{rotat}$ & learning rate for rotation & 1e-3 \\
    $\lambda_{\mathcal{D}}$ & weighting term for depth loss in mapping & 0.10 \\ 
    $\lambda_{\mathcal{S}}$ & weighting term for semantic loss in mapping & 0.10 \\ 
    $\lambda_\ell$ & re-weighting term for line feature loss & 0.25 \\
    $\mathcal{T}_{\ell}$ & line segment filtering threshold & 0.08 \\
    $\mathcal{T}_{density}$ & density threshold for hole detection & 0.90 \\
    $b_{\mathcal{G}}$ & batch size for PointNet++ \cite{qi2017pointnet++} training & 1e4 \\
    $n_{e}$ & number of epoch to train PointNet++ \cite{qi2017pointnet++} & 25 \\
    \hline
    \end{tabularx}
    \label{tab:hyperparameter}
\end{center}
\end{table}

\subsection{Evaluation on Replica-V1 Dataset}

\begin{table*}[!tp]
    \centering
    \caption{Quantitative comparison of our method and the baselines in training view rendering on the Replica-V1 dataset \cite{straub2019replica}.}
    \small
    \begin{tabularx}{\textwidth}{@{}@{\hspace{1pt}}c@{\hspace{3pt}}|>{\hspace{2pt}}l@{\hspace{1pt}} | *{11}{>{\centering\arraybackslash}X}@{}}
        \Xhline{2\arrayrulewidth}
        & \textbf{Methods} & \textbf{Metrics} & \textbf{Avg.} & Room0 & Room1 & Room2 & Office0 & Office1 & Office2 & Office3 & Office4 \\
        \hline
        && PSNR$\uparrow$ & 30.54 & 28.88 & 28.51 & 29.37 & 35.44 & 34.63 & 26.56 & 28.79 & 32.16 \\
        \multirow{0}{*}{\rotatebox[origin=c]{90}{NeRF-SLAM}} & Co-SLAM~\cite{wang2023co}~ & SSIM$\uparrow$ & 0.850 & 0.892 & 0.843 & 0.851 & 0.854 & 0.826 & 0.814 & 0.866 & 0.856 \\
        && LPIPS$\downarrow$ & 0.188 & 0.213 & 0.205 & 0.215 & 0.177 & 0.181 & 0.172 & 0.163 & 0.176 \\
        \cline{2-12}
        && PSNR$\uparrow$ & 29.08 & 25.32 & 27.77 & 29.08 & 33.71 & 30.20 & 28.09 & 28.77 & 29.71 \\
        & ESLAM~\cite{johari2023eslam}~& SSIM$\uparrow$ & 0.929 & 0.875 & 0.902 & 0.932 & 0.960 & 0.923 & 0.943 & 0.948 & 0.945 \\
        && LPIPS$\downarrow$ & 0.336 & 0.313 & 0.298 & 0.248 & 0.184 & 0.228 & 0.241 & 0.196 & 0.204 \\
        \cline{2-12}
        && PSNR$\uparrow$ & 35.17 & 32.40 & 34.08 & 35.50 & 38.26 & 39.16 & \cellcolor{thd}\textbf{33.99} & \cellcolor{thd}\textbf{33.48} & 33.49 \\
        & Point-SLAM~\cite{sandstrom2023point}~ & SSIM$\uparrow$ & \cellcolor{thd}\textbf{0.975} & \cellcolor{thd}\textbf{0.974} & \cellcolor{thd}\textbf{0.975} & \cellcolor{thd}\textbf{0.980} & \cellcolor{thd}\textbf{0.983} & \cellcolor{thd}\textbf{0.986} & \cellcolor{thd}\textbf{0.960} & 0.960 &  \cellcolor{sec}\textbf{0.979} \\
        && LPIPS$\downarrow$ & 0.124 & 0.113 & 0.116 & 0.111 & 0.100 & 0.118 & 0.156 & 0.132 & 0.142 \\
        \hline
        && PSNR$\uparrow$ & 33.98 & \cellcolor{thd}\textbf{32.48} & 33.72 & 34.96 & 38.34 & 39.04 & 31.90 & 29.70 & 31.68 \\
        & SplaTAM~\cite{keetha2023splatam}~ & SSIM$\uparrow$ & 0.969 & \cellcolor{sec}\textbf{0.975} & 0.970 & \cellcolor{fst}\textbf{0.982} & 0.982 & 0.982 & \cellcolor{thd}\textbf{0.965} & 0.950 & 0.946 \\
        && LPIPS$\downarrow$ & 0.099 & \cellcolor{thd}\textbf{0.072} & 0.096 & \cellcolor{thd}\textbf{0.074} & 0.083 & 0.093 & \cellcolor{thd}\textbf{0.100} & 0.118 & 0.155 \\
        \cline{2-12}
        && PSNR$\uparrow$ & \cellcolor{sec}\textbf{35.68} & \cellcolor{sec}\textbf{33.78} & \cellcolor{sec}\textbf{34.32} & \cellcolor{sec}\textbf{36.56} & \cellcolor{sec}\textbf{39.14} & \cellcolor{thd}\textbf{39.83} & \cellcolor{sec}\textbf{34.47} & 33.25 & 34.08 \\
        \multirow{-0.5}{*}{\rotatebox[origin=c]{90}{Gaussian-SLAM}} & MonoGS~\cite{matsuki2023gaussian}~ & SSIM$\uparrow$ & 0.962 & 0.954 & 0.957 & 0.963 & 0.972 & 0.976 & 0.962 & \cellcolor{thd}\textbf{0.960} & 0.950 \\
        && LPIPS$\downarrow$ & \cellcolor{thd}\textbf{0.087} & \cellcolor{sec}\textbf{0.071} & \cellcolor{thd}\textbf{0.086} & 0.075 & \cellcolor{thd}\textbf{0.074} & 0.087 & \cellcolor{sec}\textbf{0.098} & \cellcolor{thd}\textbf{0.098} & \cellcolor{sec}\textbf{0.105} \\
        \cline{2-12}
        && PSNR$\uparrow$ & 34.95 & 30.71 & 33.51 & 35.02 & 38.47 & 39.08 & 33.03 & \cellcolor{sec}\textbf{33.78} & \cellcolor{sec}\textbf{36.02}  \\
        & Photo-SLAM~\cite{huang2024photo}~ & SSIM$\uparrow$ & 0.942 & 0.899 & 0.934 & 0.951 & 0.964 & 0.961 & 0.938 & 0.938 & 0.952 \\
        && LPIPS$\downarrow$ & \cellcolor{fst}\textbf{0.059} & 0.075 & \cellcolor{fst}\textbf{0.057} & \cellcolor{fst}\textbf{0.043} & \cellcolor{fst}\textbf{0.050} & \cellcolor{fst}\textbf{0.047} & \cellcolor{fst}\textbf{0.077} & \cellcolor{fst}\textbf{0.066} & \cellcolor{fst}\textbf{0.054}  \\
        \cline{2-12}
        && PSNR$\uparrow$ & \cellcolor{thd}\textbf{35.43} & 31.56 & \cellcolor{thd}\textbf{34.21} & \cellcolor{thd}\textbf{35.57} & \cellcolor{thd}\textbf{39.11} & \cellcolor{sec}\textbf{40.27} & 33.54 & 32.76 & \cellcolor{fst}\textbf{36.48} \\
        & RTG-SLAM~\cite{peng2024rtg}~ & SSIM$\uparrow$ & \cellcolor{fst}\textbf{0.982} & 0.967 & \cellcolor{fst}\textbf{0.979} & \cellcolor{sec}\textbf{0.981} & \cellcolor{fst}\textbf{0.990} & \cellcolor{fst}\textbf{0.992} & \cellcolor{fst}\textbf{0.981} & \cellcolor{fst}\textbf{0.981} & \cellcolor{fst}\textbf{0.984} \\
        && LPIPS$\downarrow$ & 0.109 & 0.131 & 0.105 & 0.115 & \cellcolor{sec}\textbf{0.068} & \cellcolor{sec}\textbf{0.075} & 0.134 & 0.128 & 0.117 \\
        \cline{2-12}
        \noalign{\vskip 0.2pt}
        && PSNR$\uparrow$ & \cellcolor{fst}\textbf{36.90} & \cellcolor{fst}\textbf{34.67} &
        \cellcolor{fst}\textbf{35.52} &
        \cellcolor{fst}\textbf{37.10} &
        \cellcolor{fst}\textbf{40.04} &
        \cellcolor{fst}\textbf{41.38} &
        \cellcolor{fst}\textbf{35.91} &
        \cellcolor{fst}\textbf{34.85} &
        \cellcolor{thd}\textbf{35.75} \\
        & \textbf{Ours} & SSIM$\uparrow$ &
        \cellcolor{sec}\textbf{0.981} &
        \cellcolor{fst}\textbf{0.976} &
        \cellcolor{sec}\textbf{0.978} &
        \cellcolor{thd}\textbf{0.980} &
        \cellcolor{sec}\textbf{0.987} &
        \cellcolor{sec}\textbf{0.988} &
        \cellcolor{sec}\textbf{0.980} &
        \cellcolor{sec}\textbf{0.977} &
        \cellcolor{thd}\textbf{0.978} \\
        && LPIPS$\downarrow$ &
        \cellcolor{sec}\textbf{0.086} &
        \cellcolor{fst}\textbf{0.070} &
        \cellcolor{sec}\textbf{0.084} &
        \cellcolor{sec}\textbf{0.070} &
        0.076 &
        \cellcolor{thd}\textbf{0.083} &
        0.101 &
        \cellcolor{sec}\textbf{0.095} &
        \cellcolor{thd}\textbf{0.112} \\
        \Xhline{2\arrayrulewidth}
    \end{tabularx}
    \label{tab:replica_psnr}
\end{table*}

\begin{table}[!tp]
    \centering
    \caption{System comparison in terms of tracking, mapping, rendering FPS, and memory usage between our method and the neural dense baselines on the Replica dataset \cite{straub2019replica}. The values represent the average outcomes across 8 scenes. Note that RTG-SLAM \cite{peng2024rtg} and Photo-SLAM \cite{huang2024photo} incorporate ORB-SLAM2 \cite{mur2017orb} and ORB-SLAM3 \cite{campos2021orb} for tracking.}
    \footnotesize
    \setlength{\tabcolsep}{0pt}
    \begin{tabularx}{0.46\textwidth}{@{}@{\hspace{1pt}}c@{\hspace{3pt}}|>{\hspace{2pt}}l@{\hspace{3pt}}| >{\hspace{2pt}}*{6}{>{\centering\arraybackslash}X}}
    \Xhline{2\arrayrulewidth}
    & \textbf{Methods} & \makecell{Track.\\ATE\\{[cm]$\downarrow$}} & \makecell{Track.\\FPS\\{[f/s]$\uparrow$}} & \makecell{Map.\\FPS\\ {[f/s]$\uparrow$}} & \makecell{Render.\\FPS\\{[f/s]$\uparrow$}} & \makecell{SLAM\\FPS\\ {[f/s]$\uparrow$}} & \makecell{Param.\\Size\\{[mb]$\downarrow$}} \\ 
    \hline
    \multirow{-0.6}{*}{\rotatebox{90}{NeRF}} & Co-SLAM~\cite{wang2023co} & 1.12 & 10.2 & \cellcolor{thd}\textbf{10.0} & 35 & \cellcolor{thd}\textbf{9.26} & \cellcolor{fst}\textbf{8.2} \\
    & ESLAM~\cite{johari2023eslam} & 0.63 & 9.92 & 2.23 & 2.5 & 1.82 & 73.7 \\
    & Point-SLAM~\cite{sandstrom2023point} & \cellcolor{thd}\textbf{0.54} & 0.42 & 0.06 & 1.3 & 0.05 & 69.6 \\
    \hline
    & SplaTAM~\cite{keetha2023splatam} & 0.55 & 0.35 & 0.22 & 122 & 0.14 & 277.1 \\
    \multirow{-0.8}{*}{\rotatebox{90}{3DGS}} & MonoGS~\cite{matsuki2023gaussian} & 0.58 & 2.57 & 3.85 & \cellcolor{thd}\textbf{642} & 1.80 & \cellcolor{sec}\textbf{42.2} \\
    & Photo-SLAM~\cite{huang2024photo} & 0.59 & \cellcolor{sec}\textbf{38.72} & \cellcolor{fst}\textbf{30.04} & \cellcolor{fst}\textbf{765} & \cellcolor{fst}\textbf{16.46} & \cellcolor{thd}\textbf{59} \\
    & RTG-SLAM~\cite{peng2024rtg} & \cellcolor{sec}\textbf{0.49} & \cellcolor{fst}\textbf{45.36} & \cellcolor{sec}\textbf{18.67} & \cellcolor{sec}\textbf{692} & \cellcolor{sec}\textbf{12.65} & 71 \\
    & \textbf{Ours} & \cellcolor{fst}\textbf{0.45} & \cellcolor{thd}\textbf{35.32} & 3.96 & 324 & 3.76 & 78.5 \\
    \Xhline{2\arrayrulewidth}
    \end{tabularx}
    \label{tab:speed_and_memory}
\end{table}

In line with conventional evaluations from previous studies, we performed quantitative comparisons of training view rendering on the Replica-V1 dataset \cite{straub2019replica}, which comprises 8 single-room scenes. As presented in \Cref{tab:replica_psnr}, our method demonstrates superior rendering quality. This improvement stems from the use of a denser primitive distribution, as adopted in MonoGS \cite{matsuki2023gaussian}, leading to higher PSNR values, and the integration of the segment reconstruction loss, which contributes to a competitive structural similarity score.

\begin{table}[!tp]
    \centering
    \caption{Quantitative comparison of our method and the baseline approaches in training view rendering on the Replica Apartment dataset \cite{straub2019replica}. The underline indicates that relocalization was triggered due to accumulated tracking errors. The dash indicates system failures.}
    \setlength{\tabcolsep}{0pt} 
    \footnotesize 
    \begin{tabularx}{0.46\textwidth}{@{}@{\hspace{1pt}}c@{\hspace{3pt}}|>{\hspace{2pt}}l@{\hspace{1pt}} | >{\hspace{2pt}}c@{\hspace{1pt}} *{6}{>{\centering\arraybackslash}X}@{}}
        \Xhline{2\arrayrulewidth}
        & \textbf{Methods} & \textbf{Metrics} & \textbf{Avg.} & ap.0 & ap.1 & ap.2 & frl.0 & frl.4 \\
        \hline
        && PSNR$\uparrow$ & 24.82 & \cellcolor{thd}\textbf{29.10} & 22.86 & 23.29 & 23.52 & 25.33 \\
        \multirow{0}{*}{\rotatebox[origin=c]{90}{NeRF-SLAM}} & Co-SLAM~\cite{wang2023co}~ & SSIM$\uparrow$ & 0.816 & 0.905 & 0.766 & 0.771 & 0.822 & 0.814 \\
        && LPIPS$\downarrow$ & 0.410 & 0.321 & 0.440 & 0.462 & 0.367 & 0.461 \\
        \cline{2-9}
        && PSNR$\uparrow$ & 24.05 & 25.34 & 21.75 & 22.64 & 24.63 & 25.90 \\
        & ESLAM~\cite{johari2023eslam}~ & SSIM$\uparrow$ & 0.818 & 0.866 & 0.752 & 0.794 & 0.837 & 0.842 \\
        && LPIPS$\downarrow$ & 0.350 & 0.375 & 0.392 & 0.351 & 0.327 & 0.305 \\
        \cline{2-9}
        && PSNR$\uparrow$ & \cellcolor{sec}\textbf{34.28} & \cellcolor{fst}\textbf{34.95} & \cellcolor{sec}\textbf{32.27} & \cellcolor{fst}\textbf{33.31} & \cellcolor{sec}\textbf{36.01} & \cellcolor{sec}\textbf{34.87} \\
        & Point-SLAM~\cite{sandstrom2023point}~ & SSIM$\uparrow$ & \cellcolor{sec}\textbf{0.955} & \cellcolor{fst}\textbf{0.972} & \cellcolor{sec}\textbf{0.929} & \cellcolor{sec}\textbf{0.944} & \cellcolor{thd}\textbf{0.960} & \cellcolor{sec}\textbf{0.970} \\
        && LPIPS$\downarrow$ & \cellcolor{fst}\textbf{0.180} & \cellcolor{fst}\textbf{0.153} & \cellcolor{fst}\textbf{0.205} & \cellcolor{fst}\textbf{0.211} & \cellcolor{sec}\textbf{0.156} & \cellcolor{sec}\textbf{0.176} \\
        \hline
        && PSNR$\uparrow$ & 25.16 & 13.12 & 24.57 & 25.52 & 30.72 & 31.86 \\
        & SplaTAM~\cite{keetha2023splatam}~ & SSIM$\uparrow$ & 0.790 & 0.415 & 0.821 & 0.858 & 0.924 & 0.930 \\
        && LPIPS$\downarrow$ & 0.323 & 0.656 & 0.302 & 0.258 & 0.201 & 0.198 \\
        \cline{2-9}
        && PSNR$\uparrow$ & 26.87 & 21.89 & 26.87 & 27.92 & 30.70 & 26.97 \\
        \multirow{-0.5}{*}{\rotatebox[origin=c]{90}{Gaussian-SLAM}} & MonoGS~\cite{matsuki2023gaussian}~ & SSIM$\uparrow$ & 0.868 & 0.864 & 0.856 & 0.873 & 0.886 & 0.863 \\
        && LPIPS$\downarrow$ & 0.285 & 0.397 & 0.285 & 0.273 & 0.225 & 0.247 \\
        \cline{2-9}
        && PSNR$\uparrow$ & \cellcolor{thd}\textbf{28.78} & \underline{29.07} & \underline{22.73} & 24.59 & \cellcolor{thd}\textbf{34.16} & \cellcolor{thd}\textbf{33.36} \\
        & Photo-SLAM~\cite{huang2024photo}~ & SSIM$\uparrow$ & 0.888 & \cellcolor{thd}\textbf{\underline{0.922}} & \underline{0.796} & 0.848 & \cellcolor{thd}\textbf{0.940} & \cellcolor{thd}\textbf{0.932} \\
        && LPIPS$\downarrow$ & \cellcolor{thd}\textbf{0.224} & \cellcolor{sec}\textbf{\underline{0.227}} & \underline{0.293} & 0.354 & \cellcolor{fst}\textbf{0.115} & \cellcolor{fst}\textbf{0.129} \\
        \cline{2-9}
        && PSNR$\uparrow$ & - & - & \cellcolor{thd}\textbf{29.08} & \cellcolor{thd}\textbf{29.14} & 33.88 & - \\
        & RTG-SLAM~\cite{peng2024rtg}~ & SSIM$\uparrow$ & - & - & \cellcolor{thd}\textbf{0.900} & \cellcolor{thd}\textbf{0.909} & 0.933 & - \\
        && LPIPS$\downarrow$ & - & - & \cellcolor{sec}\textbf{0.232} & \cellcolor{sec}\textbf{0.232} & 0.181 & -  \\
        \cline{2-9}
        \noalign{\vskip 0.2pt}
        && PSNR$\uparrow$ &
        \cellcolor{fst}\textbf{34.35} & \cellcolor{sec}\textbf{34.54} &
        \cellcolor{fst}\textbf{32.37} &
        \cellcolor{sec}\textbf{32.92} &
        \cellcolor{fst}\textbf{36.24} &
        \cellcolor{fst}\textbf{35.66} \\
        & \textbf{Ours} & SSIM$\uparrow$ &
        \cellcolor{fst}\textbf{0.956} &
        \cellcolor{sec}\textbf{0.970} &
        \cellcolor{sec}\textbf{0.925} &
        \cellcolor{fst}\textbf{0.946} &
        \cellcolor{fst}\textbf{0.966} &
        \cellcolor{fst}\textbf{0.972} \\
        && LPIPS$\downarrow$ &
        \cellcolor{sec}\textbf{0.223} &
        \cellcolor{thd}\textbf{0.269} &
        \cellcolor{thd}\textbf{0.255} &
        \cellcolor{thd}\textbf{0.244} &
        \cellcolor{thd}\textbf{0.168} &
        \cellcolor{thd}\textbf{0.180} \\
        \Xhline{2\arrayrulewidth}
    \end{tabularx}
    \label{tab:apartment_psnr}
\end{table}

Additionally, \Cref{tab:speed_and_memory} provides comparisons of tracking accuracy and system efficiency. By leveraging line features, our approach achieves state-of-the-art tracking performance compared to baseline methods, further contributing to optimal map reconstruction quality. Regarding system efficiency, we assessed the frame rate of each component and memory usage. Unlike systems employing frame-to-frame tracking \cite{wang2023co,johari2023eslam,sandstrom2023point,keetha2023splatam,matsuki2023gaussian}, which exhibit significantly low tracking speeds, methods such as Photo-SLAM \cite{huang2024photo}, RTG-SLAM \cite{peng2024rtg}, and ours achieve notable advantages in tracking performance with exceptional frame rates. This is attributed to the incorporation of traditional feature-based optimization, which avoids iterative rendering and photometric loss computations for camera pose optimization—a common bottleneck in neural dense systems—thereby remarkably enhancing tracking speed. For mapping, compared to Photo-SLAM \cite{huang2024photo} and RTG-SLAM \cite{peng2024rtg}, which utilize relatively sparse primitive representations, our method opts for a denser map to improve reconstruction quality. While this results in slower mapping speeds and moderate memory usage, it offers a balanced trade-off for higher reconstruction fidelity.

\begin{figure*}[!tbp]
    \begin{center}
        \includegraphics[width=\textwidth]{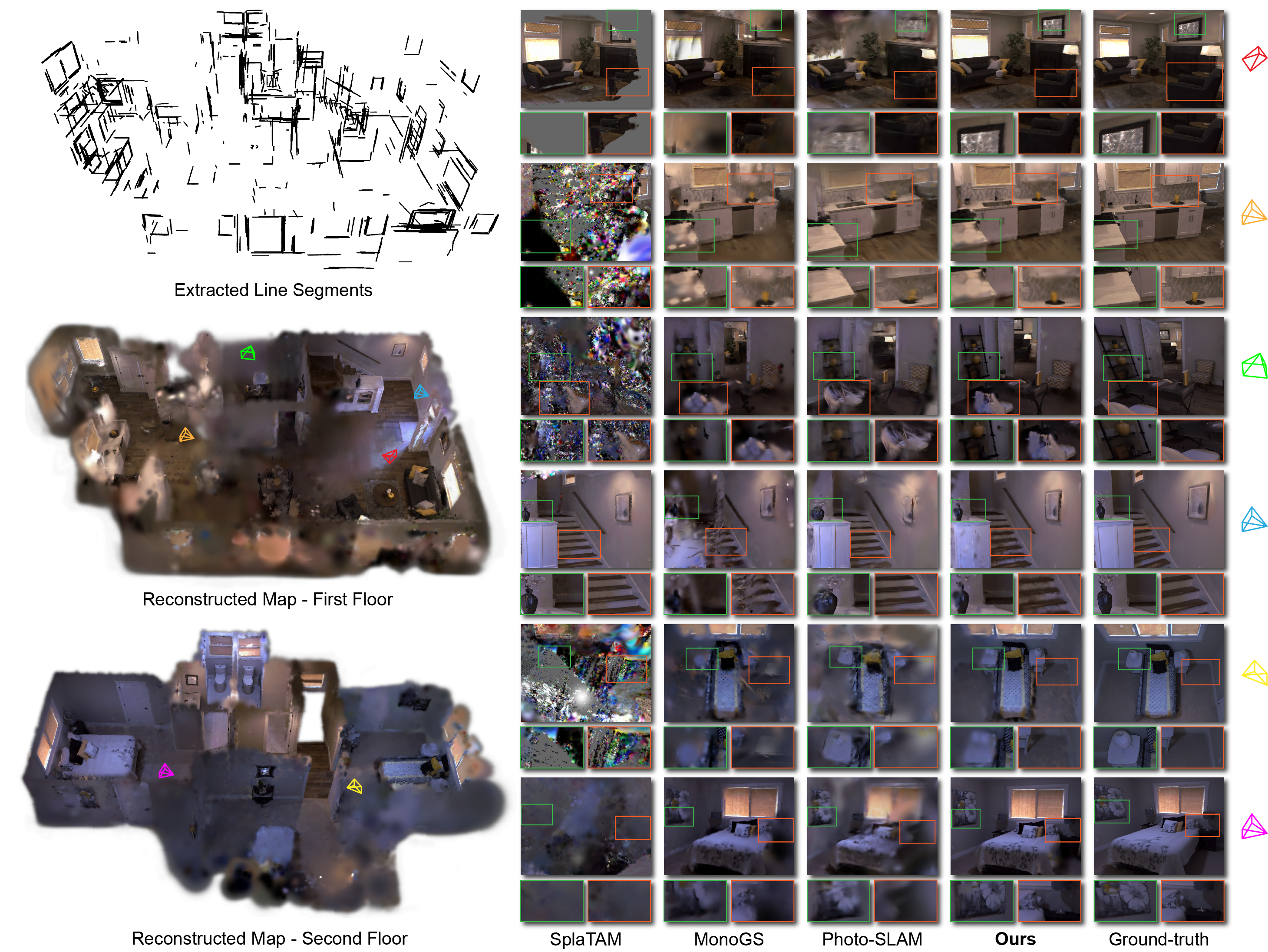}
    \end{center}
    \caption{The novel view sythesis of the scene \textit{apartment\_0} from the Replica Apartment dataset \cite{straub2019replica}. The top left shows the line segments extracted in 3D space. The bottom left illustrates the overall reconstructed scene.}
    \label{fig:apartment0}
\end{figure*}

\subsection{Evaluation on Replica-Apartment Dataset}

\begin{table}[!tp]
    \centering
    \caption{Quantitative comparison of our method with traditional, NeRF-based, and Gaussian-based RGB-D SLAM systems in terms of ATE [cm] on the Replica Apartment dataset \cite{straub2019replica}. The underlined values indicate that relocalization was necessary due to loss of tracking.}
    \footnotesize
    \begin{tabularx}{0.46\textwidth}{@{}l@{\hspace{3pt}} | >{\hspace{0pt}}l@{\hspace{3pt}} | *{7}{>{\centering\arraybackslash}X}@{}}
        \hline
        & \textbf{Methods} & \textbf{Avg.} & ap.0 & ap.1 & ap.2 & frl.0 & frl.4 \\
        \hline
        \multirow{-0.5}{*}{\rotatebox{90}{Trad.}} & ORB-SLAM2~\cite{mur2017orb} & 5.06 & \underline{6.85} & 9.83 & 6.71 &  \cellcolor{thd}\textbf{1.29} &  \cellcolor{thd}\textbf{1.00} \\
        & ORB-SLAM3~\cite{campos2021orb} & 4.41 &  \cellcolor{sec}\textbf{\underline{5.39}} & 7.98 & 6.31 &  \cellcolor{sec}\textbf{1.09} & 1.27 \\
        & PLVS~\cite{freda2023plvs} &  \cellcolor{sec}\textbf{3.72} & 7.06 &  \cellcolor{sec}\textbf{4.67} &  \cellcolor{thd}\textbf{4.04} & 1.45 & 1.39 \\
        \hline
        \multirow{-0.6}{*}{\rotatebox{90}{NeRF}} & Co-SLAM~\cite{wang2023co} &  \cellcolor{thd}\textbf{4.14} &  \cellcolor{thd}\textbf{5.30} &  \cellcolor{thd}\textbf{4.78} & 6.31 & 2.01 & 2.30 \\
        & ESLAM~\cite{johari2023eslam} & 5.54 & 9.22 & 7.09 & 7.01 & 3.33 & 1.07 \\
        & Point-SLAM~\cite{sandstrom2023point} & 5.88 & 6.39 & 17.28 &  \cellcolor{sec}\textbf{3.54} & 1.02 & 1.19 \\
        \hline
        & SplaTAM~\cite{keetha2023splatam} & 6.97 & \underline{25.14} & \underline{15.67} & 6.17 & 2.74 & 3.31 \\
        \multirow{-0.6}{*}{\rotatebox{90}{3DGS}} & MonoGS~\cite{campos2021orb} & 6.55 & 20.69 & 9.51 & 7.57 & 2.98 & 6.14 \\
        & Photo-SLAM~\cite{huang2024photo} & 4.41 &  \cellcolor{sec}\textbf{\underline{5.39}} & 7.98 & 6.31 &  \cellcolor{sec}\textbf{1.09} & 1.27 \\
        & RTG-SLAM~\cite{peng2024rtg} & 5.06 & \underline{6.99} & 9.09 & 6.90 & 1.33 &  \cellcolor{sec}\textbf{0.98} \\
        \noalign{\vskip 0.2pt}
        & \textbf{Ours} & \cellcolor{fst}\textbf{2.77} & \cellcolor{fst}\textbf{5.03} & \cellcolor{fst}\textbf{3.87} & \cellcolor{fst}\textbf{3.15} & \cellcolor{fst}\textbf{0.92} & \cellcolor{fst}\textbf{0.88} \\
        \hline
    \end{tabularx}
    \label{tab:apartment_ate}
\end{table}

To evaluate the robustness of the system in large-scale indoor environments, we evaluate MG-SLAM on the Replica Apartment dataset \cite{straub2019replica}. This dataset contains extensive multi-room scenes, complex object geometries, and looping trajectories across rooms. \Cref{tab:apartment_psnr} presents the rendering quality of our method compared to both NeRF-based and Gaussian-based approaches over five selected scenes. MG-SLAM shows notable improvements, particularly achieving a 7dB improvement in the \textit{apartment\_0} scene over Gaussian SLAM systems. This optimal performance is largely attributed to the inclusion of the fused line segments, which lays a solid foundation for loop closure and pose optimization. Compared to the NeRF-based Point-SLAM \cite{sandstrom2023point}, which utilizes additional neural point clouds for robust tracking and fine-grained mapping, our method delivers superior averaged reconstruction outcomes while maintaining optimal real-time processing capabilities, achieving a system frame rate 50 times faster.

\Cref{tab:apartment_ate} presents the quantitative evaluations of tracking ATE [cm] on the apartment scenes. Our method achieves state-of-the-art tracking accuracy compared to both frame-to-frame and feature-based tracking systems. Specifically, ORB-based systems \cite{mur2017orb, campos2021orb, huang2024photo, peng2024rtg} struggle in textureless planar environments, such as the stairs or corridors in \textit{apartmen\_0} and \textit{apartment\_1}, leading to tracking failures, frequent relocalizations, and errors in the reconstructed scenes. Leveraging the segment fusion strategy, which provides robust line features, our method also outperforms PLVS \cite{freda2023plvs} under scenarios involving extensive camera movement and rotation which are commonly encountered in the apartment dataset.



Moreover, we conducted qualitative comparisons on novel-view synthesis against Gaussian-based approaches \cite{keetha2023splatam, matsuki2023gaussian, huang2024photo} as shown in \Cref{fig:apartment0}. These comparisons highlight the superior performance of our method over baseline approaches in large-scale complex indoor environments.

\subsection{Evaluation on ScanNet Dataset}

\begin{figure*}[tb]
    \begin{center}
        \includegraphics[width=\textwidth]{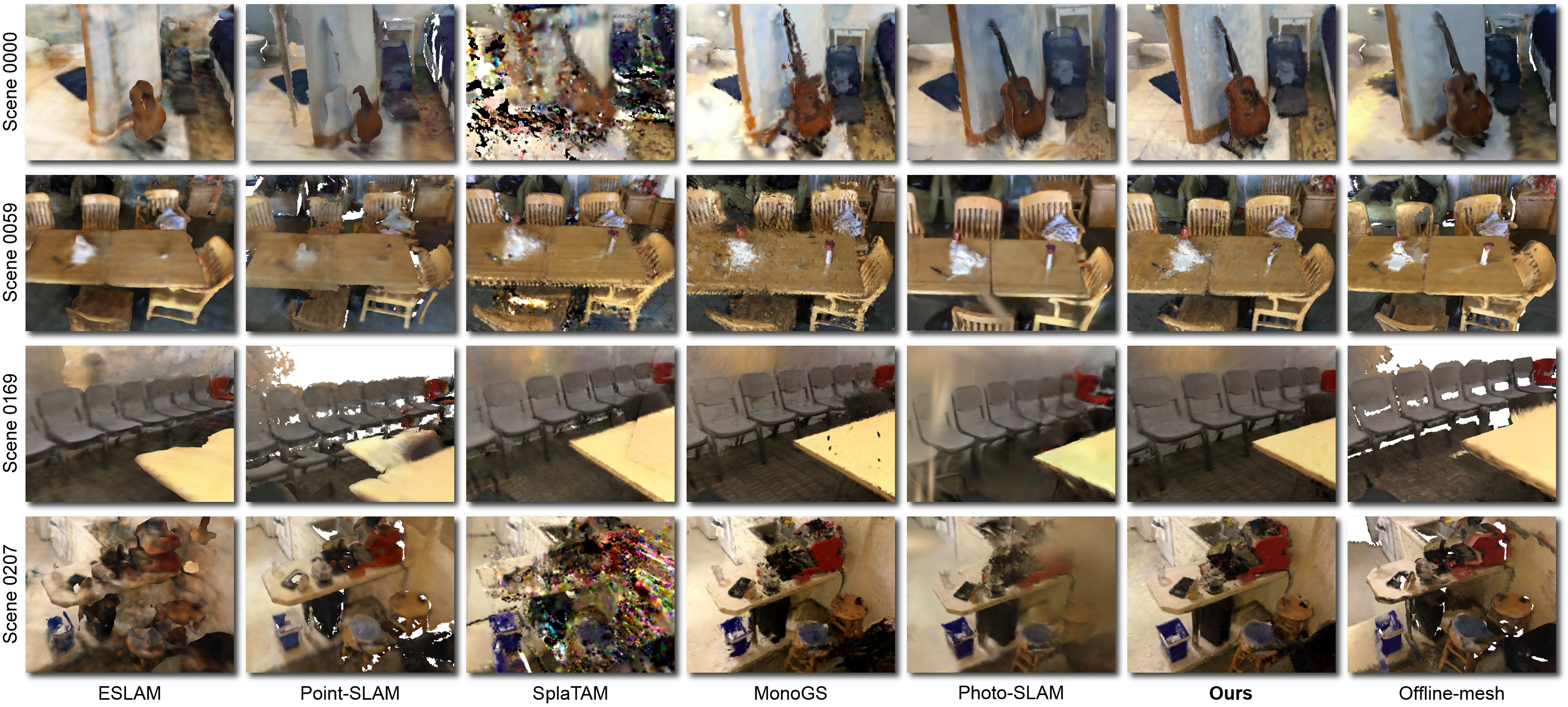}
    \end{center}
    \caption{Qualitative comparison of our method and the baselines for novel-view synthesis on the ScanNet dataset \cite{dai2017scannet}. The outcomes show that our method provides more robust and fine-grained reconstructions in real-world complex scenes compared to current NeRF-based and Gaussian-based approaches.}
    \label{fig:scannet_compare}
\end{figure*}

\begin{table}[!tp]
    \caption{Quantitative comparison of our method and the baselines in training view rendering on the ScanNet dataset \cite{dai2017scannet}.}
    \setlength{\tabcolsep}{0pt} 
    \footnotesize 
    \centering
    \begin{tabularx}{0.46\textwidth}{@{}@{\hspace{1pt}}c@{\hspace{3pt}}|>{\hspace{2pt}}l@{\hspace{1pt}} | >{\hspace{2pt}}c@{\hspace{1pt}} *{6}{>{\centering\arraybackslash}X}@{}}
        \Xhline{2\arrayrulewidth}
        & \textbf{Methods} & \textbf{Metrics} & \textbf{Avg.} & 0000 & 0059 & 0106 & 0169 & 0207 \\
        \hline
        && PSNR$\uparrow$ & 19.26 & \cellcolor{thd}\textbf{20.47} & 18.56 & 18.47 & 18.71 & 20.12 \\
        \multirow{0}{*}{\rotatebox[origin=c]{90}{NeRF-SLAM}} & Co-SLAM~\cite{wang2023co}~ & SSIM$\uparrow$ & 0.692 & 0.701 & 0.680 & 0.672 & 0.684 & 0.724 \\
        && LPIPS$\downarrow$ & 0.573 & 0.559 & 0.581 & 0.585 & 0.578 & 0.564 \\
        \cline{2-9}
        && PSNR$\uparrow$ & 16.27 & 16.67 & 15.34 & 16.52 & 15.51 & 17.32 \\
        & ESLAM~\cite{johari2023eslam}~ & SSIM$\uparrow$ & 0.650 & 0.686 & 0.632 & 0.627 & 0.655 & 0.650 \\
        && LPIPS$\downarrow$ & 0.492 & 0.452 & 0.452 & 0.528 & 0.492 & 0.538 \\
        \cline{2-9}
        && PSNR$\uparrow$ & 18.61 & 20.12 & 18.58 & 16.47 & 18.23 & 19.66 \\
        & Point-SLAM~\cite{sandstrom2023point}~ & SSIM$\uparrow$ & 0.737 & \cellcolor{sec}\textbf{0.806} & 0.765 & 0.676 & 0.686 & 0.750 \\
        && LPIPS$\downarrow$ & 0.523 & 0.485 & 0.499 & 0.544 & 0.542 & 0.544 \\
        \hline
        && PSNR$\uparrow$ & 19.02 & 17.81 & \cellcolor{thd}\textbf{19.60} & 19.23 & \cellcolor{thd}\textbf{20.55} & 17.95 \\
        & SplaTAM~\cite{keetha2023splatam}~ & SSIM$\uparrow$ & 0.726 & 0.602 & \cellcolor{thd}\textbf{0.796} & 0.741 & 0.785 & 0.705 \\
        && LPIPS$\downarrow$ & 0.337 & 0.467 & 0.290 & 0.322 & 0.260 & 0.346 \\
        \cline{2-9}
        && PSNR$\uparrow$ & \cellcolor{thd}\textbf{20.08} & 19.76 & 19.25 & \cellcolor{thd}\textbf{20.18} & \cellcolor{sec}\textbf{20.57} & \cellcolor{thd}\textbf{20.62} \\
        \multirow{-0.5}{*}{\rotatebox[origin=c]{90}{Gaussian-SLAM}} & MonoGS~\cite{matsuki2023gaussian}~ & SSIM$\uparrow$ & \cellcolor{thd}\textbf{0.782} & \cellcolor{thd}\textbf{0.772} & 0.767 & \cellcolor{thd}\textbf{0.785} & \cellcolor{thd}\textbf{0.790} & \cellcolor{sec}\textbf{0.798} \\
        && LPIPS$\downarrow$ & \cellcolor{thd}\textbf{0.300} & \cellcolor{thd}\textbf{0.387} & \cellcolor{thd}\textbf{0.289} & \cellcolor{thd}\textbf{0.272} & \cellcolor{thd}\textbf{0.256} & \cellcolor{thd}\textbf{0.295} \\
        \cline{2-9}
        && PSNR$\uparrow$ & \cellcolor{sec}\textbf{20.76} & \cellcolor{sec}\textbf{21.74} & \cellcolor{sec}\textbf{20.07} & \cellcolor{sec}\textbf{20.70} & 20.34 & \cellcolor{sec}\textbf{20.94} \\
        & Photo-SLAM~\cite{huang2024photo}~ & SSIM$\uparrow$ & \cellcolor{sec}\textbf{0.790} & 0.771 & \cellcolor{sec}\textbf{0.805} & \cellcolor{sec}\textbf{0.792} & \cellcolor{sec}\textbf{0.792} & \cellcolor{thd}\textbf{0.790} \\
        && LPIPS$\downarrow$ & \cellcolor{sec}\textbf{0.293} & \cellcolor{sec}\textbf{0.375} & \cellcolor{sec}\textbf{0.280} & \cellcolor{sec}\textbf{0.269} & \cellcolor{sec}\textbf{0.248} & \cellcolor{sec}\textbf{0.292} \\
        \cline{2-9}
        && PSNR$\uparrow$ & 16.79 & 18.62 & 15.56 & 14.97 & 18.07 & 18.52 \\
        & RTG-SLAM~\cite{peng2024rtg}~ & SSIM$\uparrow$ & 0.743 & 0.756 & 0.682 & 0.726 & 0.772 & 0.773 \\
        && LPIPS$\downarrow$ & 0.480 & 0.468 & 0.531 & 0.480 & 0.451 & 0.459 \\
        \cline{2-9}
        \noalign{\vskip 0.2pt}
        && PSNR$\uparrow$ & \cellcolor{fst}\textbf{23.71} & \cellcolor{fst}\textbf{25.69} &
        \cellcolor{fst}\textbf{23.62} &
        \cellcolor{fst}\textbf{22.95} &
        \cellcolor{fst}\textbf{22.86} &
        \cellcolor{fst}\textbf{23.42} \\
        & \textbf{Ours} & SSIM$\uparrow$ &
        \cellcolor{fst}\textbf{0.838} &
        \cellcolor{fst}\textbf{0.846} &
        \cellcolor{fst}\textbf{0.838} &
        \cellcolor{fst}\textbf{0.829} &
        \cellcolor{fst}\textbf{0.836} &
        \cellcolor{fst}\textbf{0.840} \\
        && LPIPS$\downarrow$ &
        \cellcolor{fst}\textbf{0.262} &
        \cellcolor{fst}\textbf{0.282} &
        \cellcolor{fst}\textbf{0.255} &
        \cellcolor{fst}\textbf{0.260} &
        \cellcolor{fst}\textbf{0.236} &
        \cellcolor{fst}\textbf{0.277} \\
        \Xhline{2\arrayrulewidth}
     \end{tabularx}
    \label{tab:scannet_psnr}
\end{table}

\begin{table}[!tp]
    \centering
    \caption{Quantitative comparison of our method and the neural dense baselines in terms of ATE [cm] on the ScanNet dataset \cite{dai2017scannet}.}
    \setlength{\tabcolsep}{0pt} 
    \footnotesize 
    \begin{tabularx}{0.46\textwidth}{@{}@{\hspace{1pt}}c@{\hspace{3pt}}|>{\hspace{2pt}}l@{\hspace{1pt}}|*{7}{>{\centering\arraybackslash}X}@{}}
        \Xhline{2\arrayrulewidth}
        & \textbf{Methods}  & \textbf{Avg.} & 0000 & 0059 & 0106 & 0169 & 0207 \\
        \hline
        \multirow{-0.5}{*}{\rotatebox[origin=c]{90}{NeRF}} & Co-SLAM~\cite{wang2023co}~ & \cellcolor{thd}\textbf{8.46} & 11.14 & 9.36 & \cellcolor{fst}\textbf{5.90} & \cellcolor{fst}\textbf{7.14} & 8.75 \\
        & ESLAM~\cite{johari2023eslam}~ & 7.68 & 8.47 & 8.70 & \cellcolor{sec}\textbf{7.58} & \cellcolor{thd}\textbf{7.45} & \cellcolor{sec}\textbf{6.20} \\
        & Point-SLAM~\cite{sandstrom2023point}~ & 11.68 & 10.24 & 7.81 & 8.65 & 22.16 & 9.54 \\
        \hline
        \multirow{0}{*}{\rotatebox[origin=c]{90}{Gaussian}} & SplaTAM~\cite{keetha2023splatam}~ & 12.04 & 12.83 & 10.10 & 17.72 & 12.08 & 7.46 \\
        & MonoGS~\cite{matsuki2023gaussian}~ & 13.86 & 15.94 & 13.03 & 19.44 & 10.44 & 10.46 \\
        & Photo-SLAM~\cite{huang2024photo}~ & \cellcolor{sec}\textbf{8.40} & \cellcolor{sec}\textbf{7.62} & \cellcolor{thd}\textbf{7.76} & 9.36 & 10.01 & \cellcolor{thd}\textbf{7.23} \\
        & RTG-SLAM~\cite{peng2024rtg}~ & 8.70 & \cellcolor{thd}\textbf{8.04} & \cellcolor{sec}\textbf{6.82} & 9.22 & 10.15 & 9.25 \\
        \noalign{\vskip 0.2pt}
        & \textbf{Ours} &
        \cellcolor{fst}\textbf{6.77} &
        \cellcolor{fst}\textbf{5.95} &
        \cellcolor{fst}\textbf{6.41} &
        \cellcolor{thd}\textbf{8.07} &
        \cellcolor{sec}\textbf{7.29} &
        \cellcolor{fst}\textbf{6.14} \\
        \Xhline{2\arrayrulewidth}
    \end{tabularx}
    \label{tab:scannet_ate}
\end{table}

We provide quantitative assessments of reconstruction quality using the ScanNet dataset \cite{dai2017scannet} in \Cref{tab:scannet_psnr}. Our approach delivers state-of-the-art results, outperforming other Gaussian-based methods by a notable 3dB in PSNR in real-world environments. The tracking evaluation results are shown in \Cref{tab:scannet_ate}. Our method remarkably reduces the ATE RMSE (cm) error, achieving 20\% improvements over baseline approaches.

In \Cref{fig:scannet_compare}, we visualize the novel-view synthesis results for MG-SLAM, comparing it with both NeRF-based and Gaussian-based SLAM systems. Our method exhibits robust and high-fidelity reconstruction capabilities. In comparison to Point-SLAM \cite{sandstrom2023point}, our method offers complete scene reconstructions and finer texture details, benefiting from the use of Gaussians to handle complex geometries. Among Gaussian-based methods, SplaTAM \cite{keetha2023splatam} struggles in large scenes with complex camera loops due to its lack of bundle adjustment in the tracking system. Similarly, MonoGS \cite{matsuki2023gaussian} generally delivers reliable quality but struggles with object-level reconstruction drift. Photo-SLAM \cite{huang2024photo} achieves better reconstruction results compared to other baseline approaches but suffers from floaters and artifacts that compromise its geometric accuracy. In contrast, MG-SLAM excels with robust tracking and effective bundle adjustment incorporating line segments, enabling superior detailed reconstructions even surpassing the ground-truth mesh derived from offline methods.

\subsection{Evaluation on Physical Platform}

\begin{figure*}[t]
    \begin{center}
        \includegraphics[width=\textwidth]{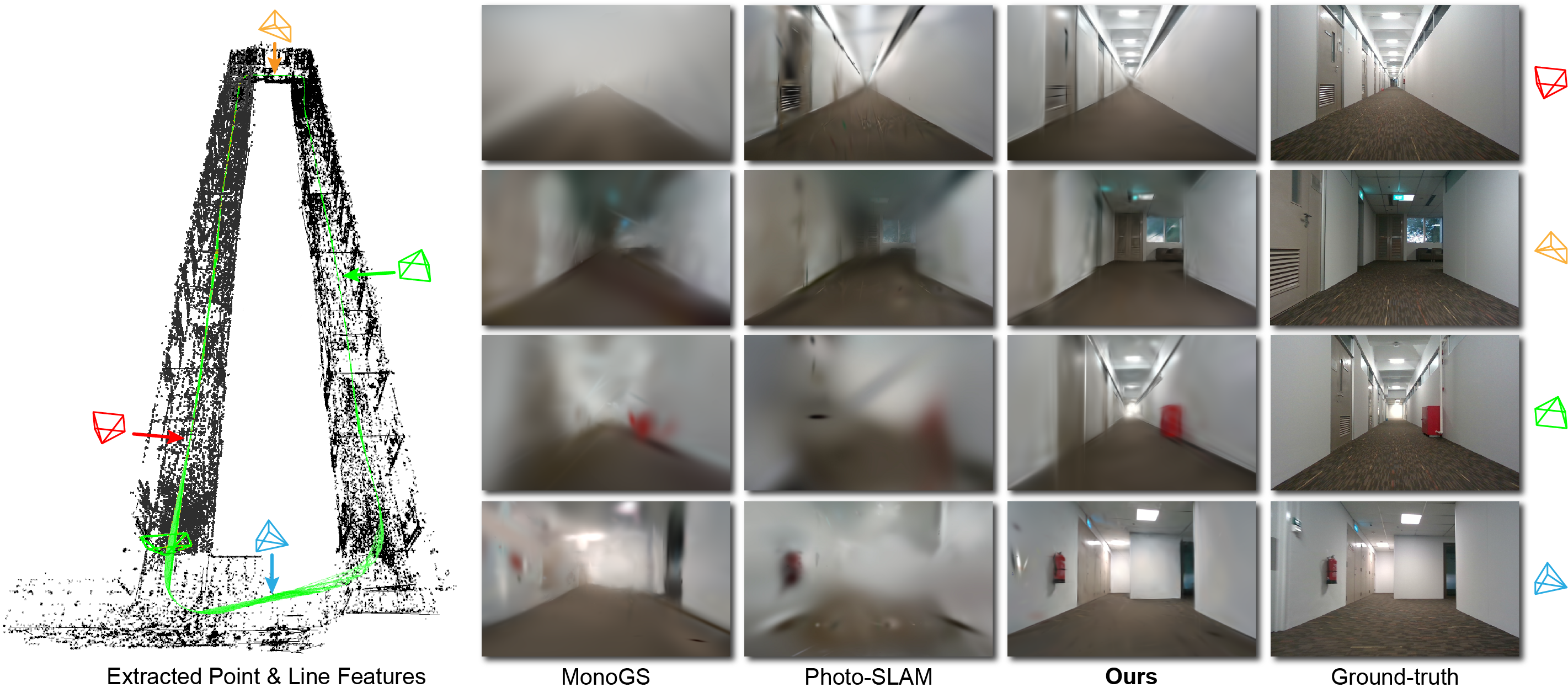}
    \end{center}
    \caption{Qualitative comparison of our method and the baselines on the trajectory collected using our physical platform. The left side displays the line and point features extracted by our tracking system. The right side presents the reconstruction comparisons with Gaussian-based methods, with MG-SLAM showing more reliable reconstruction results in long-horizon and textureless indoor environments.}
    \label{fig:ntu_compare}
\end{figure*}

\begin{table}[!tp]
    \centering
    \caption{Quantitative comparison of our method and Gaussian-based approaches in terms of tracking and mapping performance on trajectory collected using our physical platform. The dash indicates system failures.}
    \setlength{\tabcolsep}{0pt} 
    \footnotesize 
    \begin{tabularx}{0.46\textwidth}{@{}l@{\hspace{1pt}}|*{4}{>{\centering\arraybackslash}X}@{}}
        \Xhline{2\arrayrulewidth}
        \textbf{Methods}  & ATE [cm]$\downarrow$ & PSNR [dB]$\uparrow$ & SSIM$\uparrow$ & LPIPS$\downarrow$ \\
        \hline
        SplaTAM~\cite{keetha2023splatam}~ & - & - & - & - \\
        MonoGS~\cite{matsuki2023gaussian}~ & \cellcolor{thd}\textbf{27.04} & \cellcolor{thd}\textbf{18.28} & \cellcolor{thd}\textbf{0.751} & \cellcolor{thd}\textbf{0.677} \\
        Photo-SLAM~\cite{huang2024photo}~ & \cellcolor{sec}\textbf{18.29} & \cellcolor{sec}\textbf{22.92} & \cellcolor{sec}\textbf{0.798} & 
        \cellcolor{sec}\textbf{0.620} \\
        \textbf{Ours} &
        \cellcolor{fst}\textbf{7.72} &
        \cellcolor{fst}\textbf{25.47} &
        \cellcolor{fst}\textbf{0.831} &
        \cellcolor{fst}\textbf{0.575} \\
        \Xhline{2\arrayrulewidth}
    \end{tabularx}
    \label{tab:ntu_table}
\end{table}

To assess our system's performance in real-world environments, we utilzie a private data trajectory gathered with our physical platform. The robot was deployed in the NTU building to navigate and collect data from a challenging long corridor, notable for its lack of texture and extensive length exceeding 200 meters. Ground-truth poses were computed by registering the LiDAR point cloud with the scanned point cloud generated by the Leica ScanStation. \Cref{fig:ntu_compare} displays the comparisons of training view synthesis between our method and Gaussian-based systems, where our method demonstrates significant advantages; however, it still lacks some details compared to the ground-truth images due to the long-horizon trajectory. \Cref{tab:ntu_table} provides quantitative results, showing that our method achieves more accurate tracking and mapping performance. The baseline approaches struggle with tracking in textureless environments, further affirming the effectiveness of our method in indoor scenes.

\subsection{Evaluation of Scene Completion}

\begin{figure*}[!tp]
    \begin{center}
        \includegraphics[width=\textwidth]{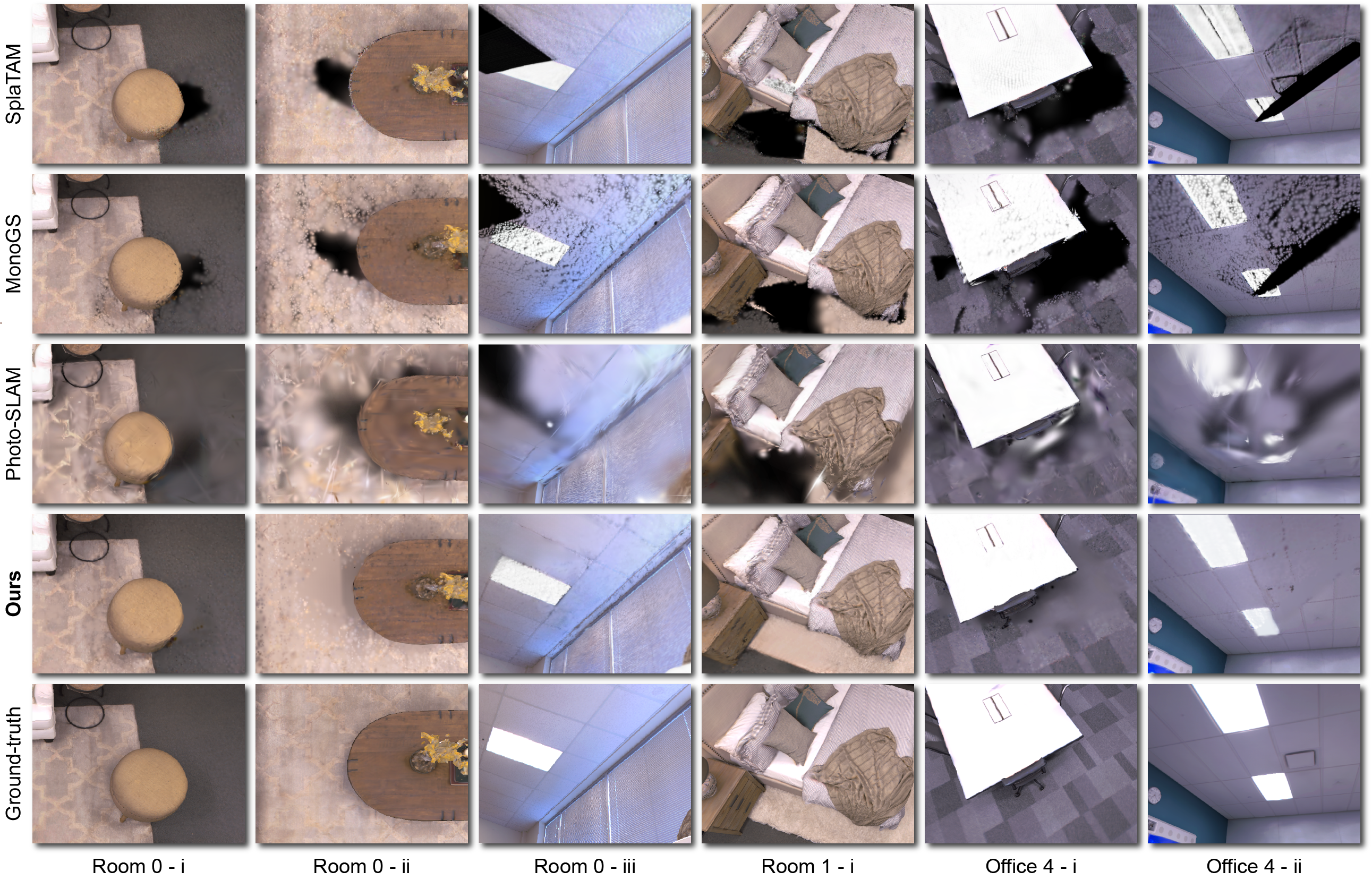}
    \end{center}
    \caption{Qualitative comparison on the novel-view synthesis of our method and the Gaussian SLAM baselines for hole fillings on the Replica dataset \cite{straub2019replica}. Our method demonstrates superior capability in interpolating and filling gaps on structured planar surfaces, such as floors and ceilings.}
    \label{fig:replica_fill_holes}
\end{figure*}

The Gaussian SLAM faces limitations in interpolating the geometry of unseen views. This issue is especially evident in indoor scenes that feature complex layouts, where basic surfaces like floors are often obscured and poorly represented. \Cref{fig:replica_fill_holes} qualitatively compared our method with recent Gaussian-based approaches in the Replica scenes \cite{straub2019replica} that presented uncovered geometry in camera trajectories. Utilizing the surface interpolation strategy based on the MW hypothesis \cite{coughlan2000manhattan}, our method can accurately detect and proactively generate new Gaussians efficiently to fill gaps with certain textures, whereas Gaussian baseline methods exhibit substantial defects on structured surfaces.

\section{Ablation Study}
\label{sec:ablation}

In this section, we provide a comprehensive ablation analysis of the hyperparameters and each component's effect.

\begin{figure}[tb]
    \begin{center}
        \includegraphics[width=0.5\textwidth]{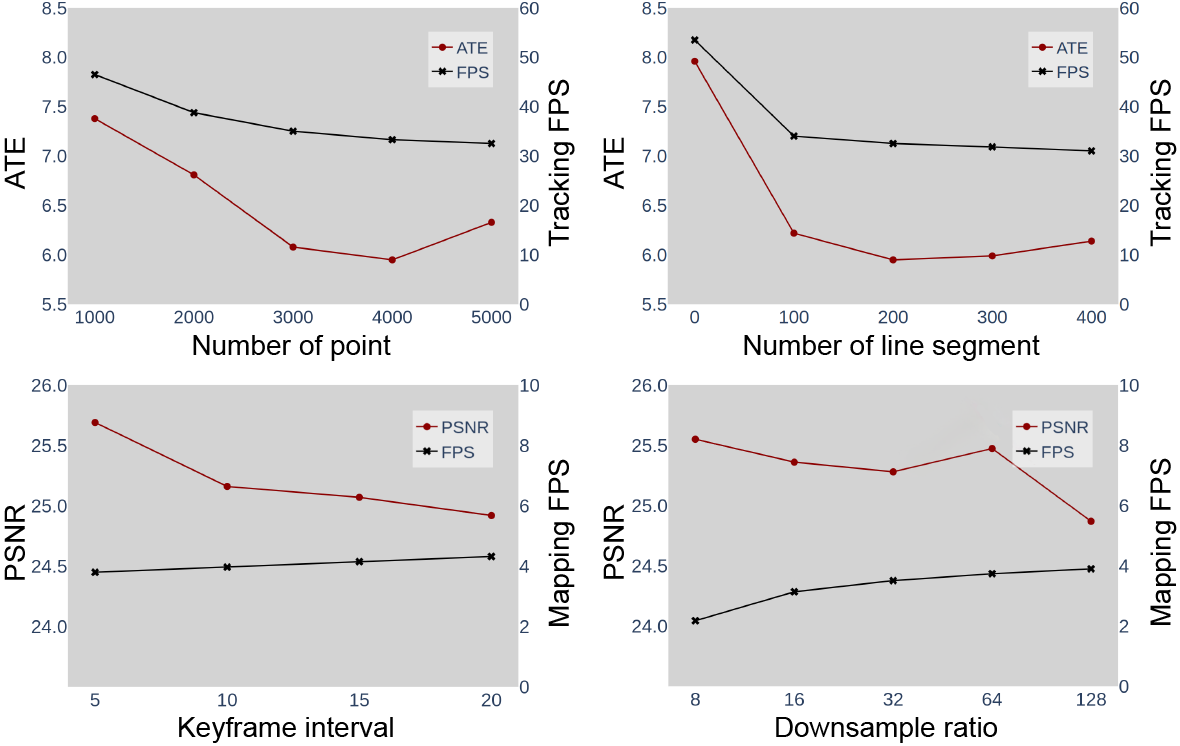}
    \end{center}
    \caption{The ablation study examining the impact of the number of points, line segments, keyframe intervals, and downsample ratios in MG-SLAM on \textit{scene0000\_00} from the ScanNet dataset \cite{dai2017scannet}.}
    \label{fig:sup_ablation_chart}
\end{figure}

\subsection{Ablation of Hyperparameters}

Our tracking system, which advances the foundation of PLVS system \cite{freda2023plvs}, utilizes feature points and fused line segments for optimizing camera poses and performing bundle adjustments. The upper row of \Cref{fig:sup_ablation_chart} illustrates the effects of varying the number of points and line segments on the ATE loss [cm] and the frame rate of the tracking system. We identified an optimal region where having too few points and lines results in a lack of sufficient anchor features, whereas too many points and lines can lead to notable feature mismatches that reduce tracking accuracy. The tracking FPS reduces as a tradeoff with the increase in the number of features. Compared to tracking solely by point features, including line segment features noticeably decreases the frame rate; however, it still maintains a high rate over 30 FPS and does not become the speed bottleneck of the overall systems. The bottom row of \Cref{fig:sup_ablation_chart} illustrates the effects of default keyframe intervals and downsample ratio on the Gaussian map. During the mapping process, we uniformly downsample the point cloud generated from RGB-D input by a specific ratio to accelerate the system and conserve memory. We observed a marginal PSNR disparity when maintaining a relatively small downsample ratio.

\subsection{Ablation of Line Segment Extraction}

\begin{figure*}[!tp]
    \begin{center}
        \includegraphics[width=0.86\textwidth]{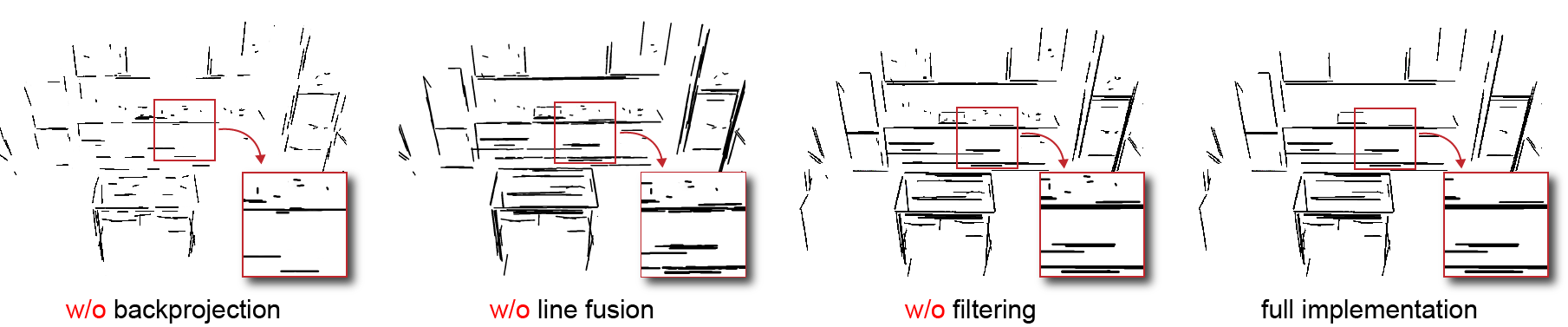}
    \end{center}
    \caption{The ablation study of the line segment extraction outcomes for our method on the scene \textit{frl\_apartment\_4} from the Replica Apartment dataset \cite{straub2019replica}. By incorporating backprojection, filtering, and segment fusion, our system generates rich and robust line segment features for subsequent tracking and mapping procedures.}
    \label{fig:line_ablation}
\end{figure*}

The extracted line segments play a crucial role in serving as robust feature foundations in the subsequent optimization processes. \Cref{fig:line_ablation} illustrates the ablation results for our feature extraction approach for these 3D line segments, demonstrating the efficacy of segment backprojection, line fusion, and filtering strategies in providing clear and accurate line features.

\begin{figure*}[!tp]
    \begin{center}
        \includegraphics[width=0.86\textwidth]{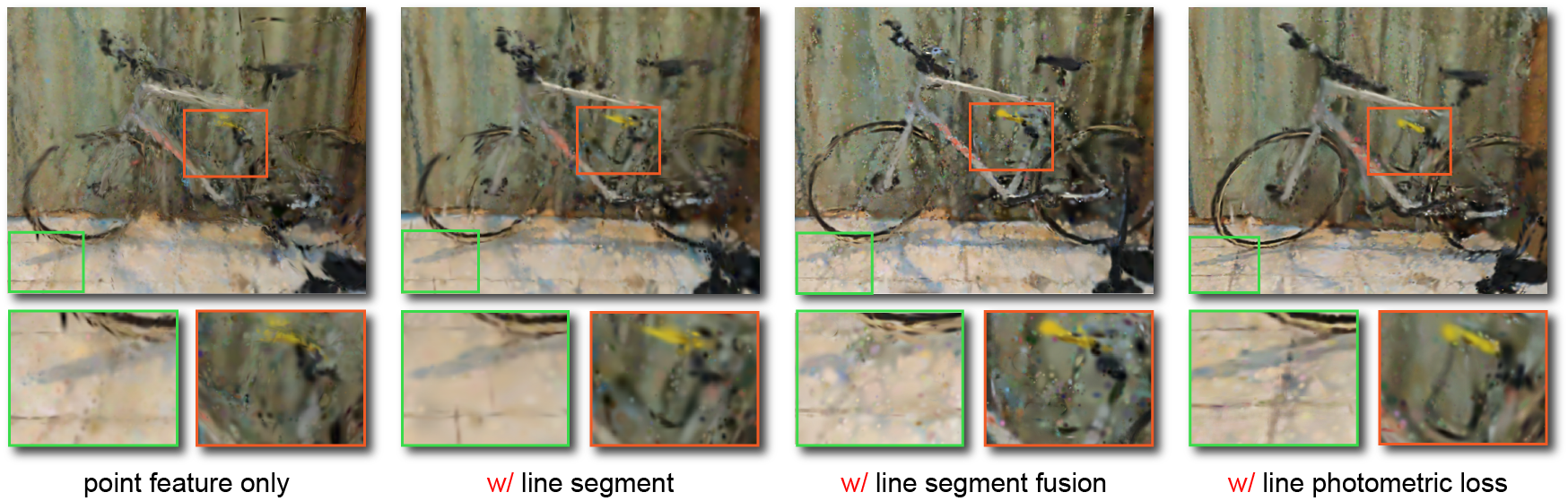}
    \end{center}
    \caption{The visualization of the ablation study in \textit{scene0000\_00} from the ScanNet dataset \cite{dai2017scannet}. From left to right, the results are displayed starting completely without line features, adding line segments, integrating segment fusion, and finally incorporating line photometric loss in mapping. Key differences are highlighted with colored boxes.}
    \label{fig:sup_ablation_fig}
\end{figure*}

\subsection{Ablation of Tracking and Mapping Loss}

\begin{table}[!tp]
    \centering
    \caption{Quantitative comparison of the ablation study on line losses, measuring ATE [cm] and PSNR [dB], conducted on \textit{scene0000\_00} of the ScanNet dataset \cite{dai2017scannet}. The checkmark symbol indicates the employment of the method.}
    \small
    \begin{tabularx}{0.46\textwidth}{*{3}{>{\centering\arraybackslash}X} || *{3}{>{\centering\arraybackslash}X}@{}}
        \Xhline{2\arrayrulewidth}
        \multicolumn{3}{c||}{\textbf{Methods}} & \multicolumn{3}{c}{\textbf{Metric}} \\
        \cline{1-3} \cline{4-6}
        line segment & segment fusion & line photo. loss & ATE [cm] & PSNR [dB] \\
        \hline
        &  &  & 7.41 & 20.65 \\
        \usym{1F5F8} &  &  & 6.58 & 22.89 \\
        \usym{1F5F8} & \usym{1F5F8} &  & \textbf{5.95} & 24.72 \\
        \usym{1F5F8} & \usym{1F5F8} & \usym{1F5F8} & \textbf{5.95} & \textbf{25.69} \\
        \hline
    \end{tabularx}
    \label{tab:ablation_line_loss}
\end{table}

MG-SLAM employs line segments for robust camera pose optimization and fine-grained map reconstruction. Specifically, fused line segments are utilized in the bundle adjustment of the tracking procedure, and a photometric loss related to line features is integrated into the map optimization. We use \textit{scene0000\_00} from the ScanNet dataset \cite{dai2017scannet} to conduct the ablation study that evaluates the impact of each loss. This scene was chosen because it contains rich line features, which allow for a clear assessment of each component's effectiveness. \Cref{tab:ablation_line_loss} presents the ATE [cm] and PSNR [dB] in relation to the use of each loss. We observed that integrating line features remarkably enhances tracking accuracy, and the fusion of line segments, which facilitates robust edge extraction, further improves the performance. For mapping, the photometric loss provides additional geometric constraints, thereby offering optimal reconstruction quality.

\begin{figure}[!tbp]
    \begin{center}
        \includegraphics[width=0.49\textwidth]{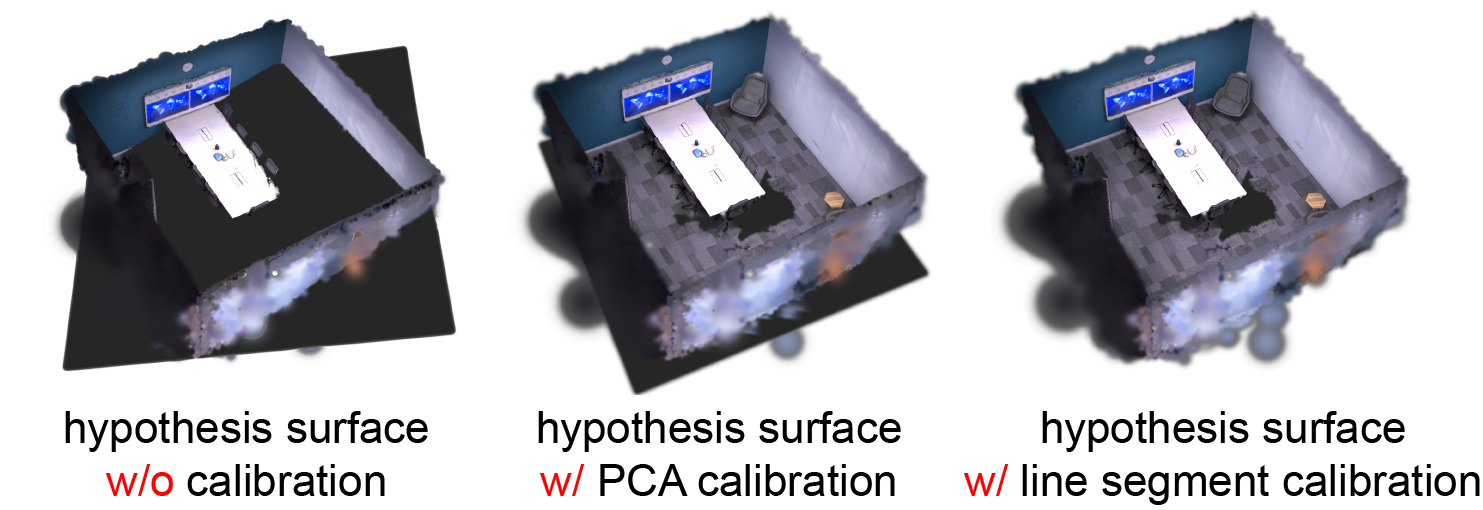}
    \end{center}
    \caption{The ablation study compares surface identification approaches. The identified hypothesis surfaces using different methods are shaded in black. The left figure displays the scene without calibration, where the primary axes of the scene do not align with the world coordinates. The middle figure demonstrates the application of PCA to the centers of Gaussians $\mu_{\mathcal{G}}$ associated with structured surfaces, where the boundaries of $\mu_{\mathcal{G}}$ define the surface. The right figure shows the calibration results, achieved by clustering line segments and defining surfaces based on the boundaries of these line features.}
    \label{fig:ablation_fillholes}
\end{figure}

In \Cref{fig:sup_ablation_fig}, we present a typical case in \textit{scene0000\_00} to demonstrate the effectiveness of each loss component. This case features a bike that appears multiple times along the camera trajectory. Compared to relying solely on point features, adding fused line segments significantly improves the quality of map reconstruction. This enhancement mitigates scene drift by providing more accurate camera pose estimation. Moreover, incorporating the photometric loss of line features in mapping results in more detailed object-level geometry and more precise reconstruction of line-like textures.

\subsection{Ablation of Surface Extraction}

We employ hypothesis surfaces regularized by line segments to pinpoint potential gaps on structured surfaces, as described in \Cref{sec:structure_opt}. These identified hypothesis surfaces are designed to precisely mirror the positions of the planar surfaces with minimal redundancy. To delineate these surfaces, we initially adjust the reconstructed map to adhere to the orthogonality assumption of MW. \Cref{fig:ablation_fillholes} displays the hypothesized rectangular planes produced without calibration, with PCA calibration based on the centers of Gaussian primitives, and with calibration using features from extracted segments. It is evident that our segment-based method efficiently captures the floor surface with reduced redundancy.

\begin{figure*}[htbp]
    \begin{center}
        \includegraphics[width=\textwidth]{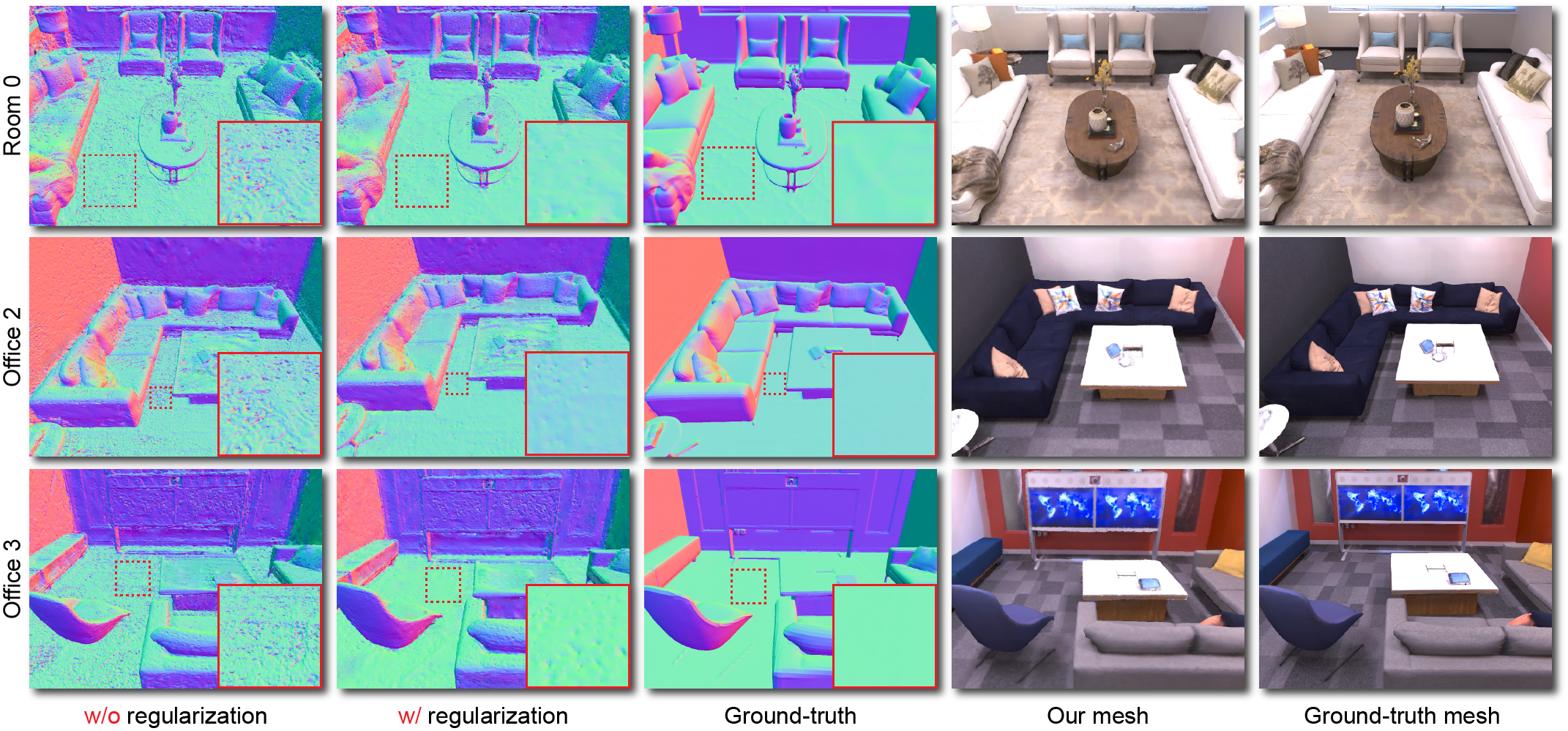}
    \end{center}
    \caption{The ablation of normal regularization for surface extraction on the Replica dataset \cite{straub2019replica}. We compare the normal map of the extracted mesh, both without and with normal regularization, to the ground-truth mesh. The regularization refines the planar surfaces of the scene, i.e. floors, effectively.}
    \label{fig:sup_normal}
\end{figure*}

\subsection{Ablation of Mesh Generation}

Following map reconstruction and surface interpolation, we extract meshes from the Gaussian map, as outlined in \Cref{sec:mesh_generate}. \Cref{fig:sup_normal} illustrates the results of mesh extraction with and without the proposed normal regularization loss. We observe that incorperating this regularization term results in smoother mesh extraction on the structured surface such as floors compared to the original method introduced by \cite{guedon2023sugar}.

\section{Conclusion}
In this study, we present MG-SLAM, a Gaussian-based SLAM method based on the MW hypothesis. MG-SLAM employs fused line segments and point features for robust pose estimation and map refinement. Furthermore, by leveraging the line segments and planar surface assumption, we efficiently generate new Gaussian primitives in gaps on structural surfaces caused by obstructions or unseen geometry. Extensive experiments demonstrate that our method delivers state-of-the-art tracking and mapping performance, while also maintaining real-time processing speed.

\section{Limitations}

MW assumes surface planes generally align with the three orthogonal directions defined by the Cartesian coordinate system. This presents challenges in refining planes or layouts that are slanted or not strictly orthogonal. To better accommodate diverse outdoor urban environments, future research could incorporate multiple horizontal dominant directions, as seen in Atlanta World \cite{schindler2004atlanta}, include additional sloping directions like those in Hong Kong World \cite{li2023hong}, or adopt uniform inclination angles for the dominant directions proposed in San Francisco World \cite{ham2024san}. Additionally, since the optimization of our dense map primarily relies on line features and large planar surfaces, it struggles with refining piece-wise or fine-grained structural planes, such as walls and windows. One potential approach is to leverage the coplanarity of surfaces for structural regularization \cite{li2021rgb, hong2024liv}, although this proves particularly challenging in indoor environments with complex layouts where identifying coplanar surfaces is difficult.




\bibliographystyle{IEEEtran}
\bibliography{ieee}

\section{Biography Section}

\begin{IEEEbiography}[{\includegraphics[width=1in,height=1.25in,clip,keepaspectratio]{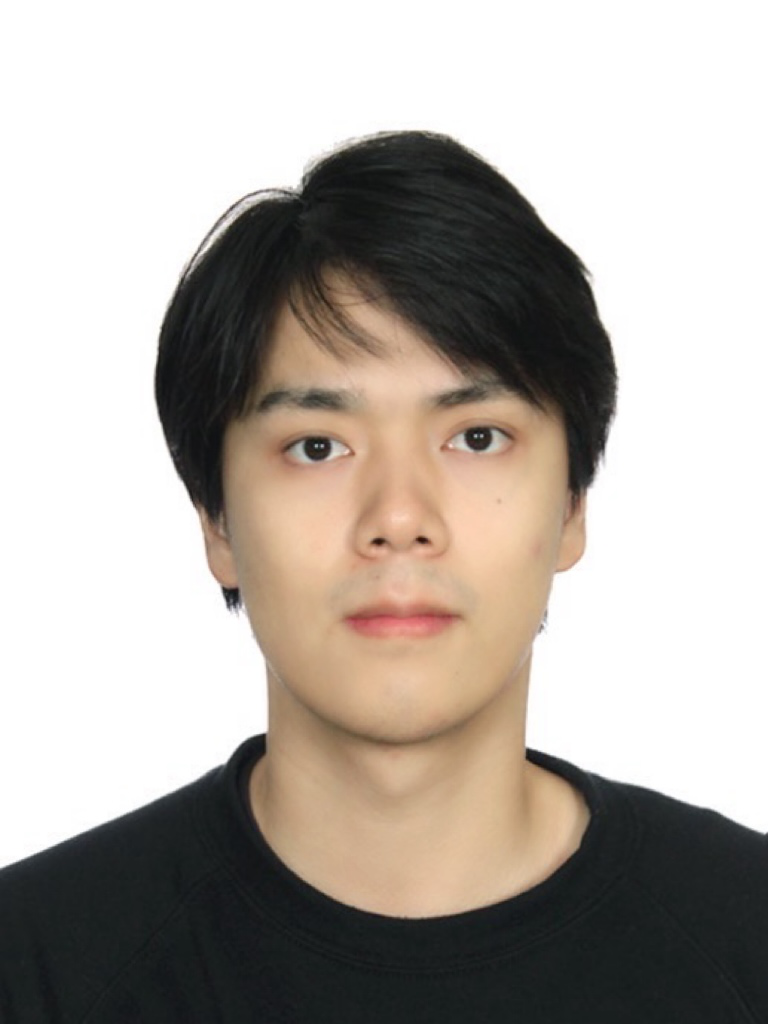}}]
{Shuhong Liu} is currently pursuing the Ph.D. degree in Department of Mechano-informatics, Information Science and Technology with the University of Tokyo, Tokyo, Japan. Before that, he received his bachelor's degree in Department of Electrical and Computer Engineering at University of Waterloo, Ontario, Canada, and the master's degree in Creative Informatics, Information Science and Technology, the University of Tokyo, Tokyo, Japan. His research interests include 3D computer vision, visual SLAM, and computation photography.
\end{IEEEbiography}

\begin{IEEEbiography}[{\includegraphics[width=1in,height=1.25in,clip,keepaspectratio]{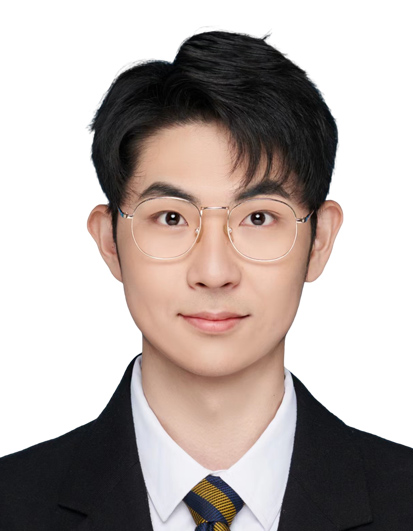}}]
{Tianchen Deng} is currently pursuing the
Ph.D. degree in control science and engineering with
Shanghai Jiao Tong University, Shanghai, China.
His main research interests include 3D Reconstruction, long-term
visual simultaneous localization and mapping
(SLAM), and vision-based localization. \end{IEEEbiography}

\begin{IEEEbiography}[{\includegraphics[width=1in,height=1.25in,clip,keepaspectratio]{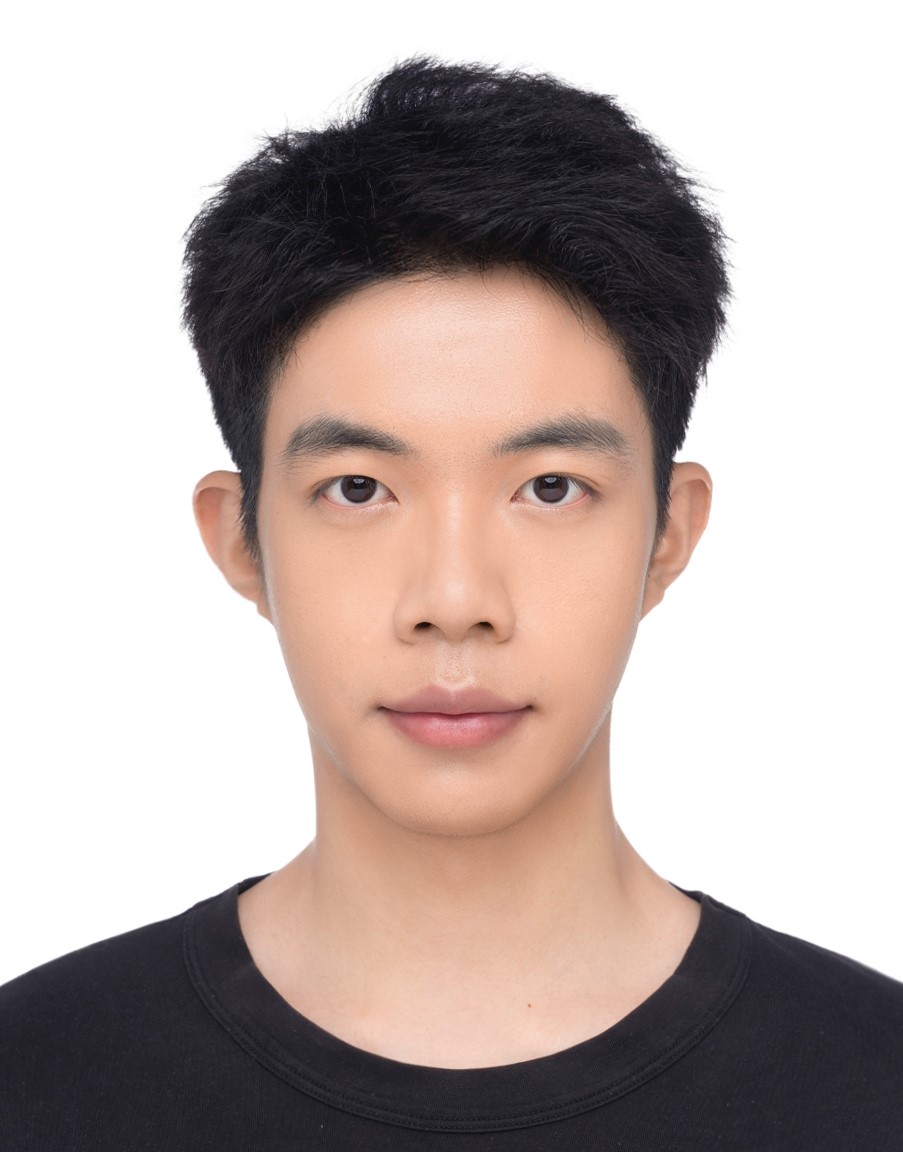}}]
{Heng Zhou} is currently a master student in Mechanical Engineering at Columbia University. He is also a member of the Creative Machines Lab, led by Professor Hod Lipson. His research interests include SLAM and 3D reconstruction. His current work focuses primarily on the integration of 3D Gaussian techniques with SLAM applications.
\end{IEEEbiography}

\begin{IEEEbiography}[{\includegraphics[width=1in,height=1.25in,clip,keepaspectratio]{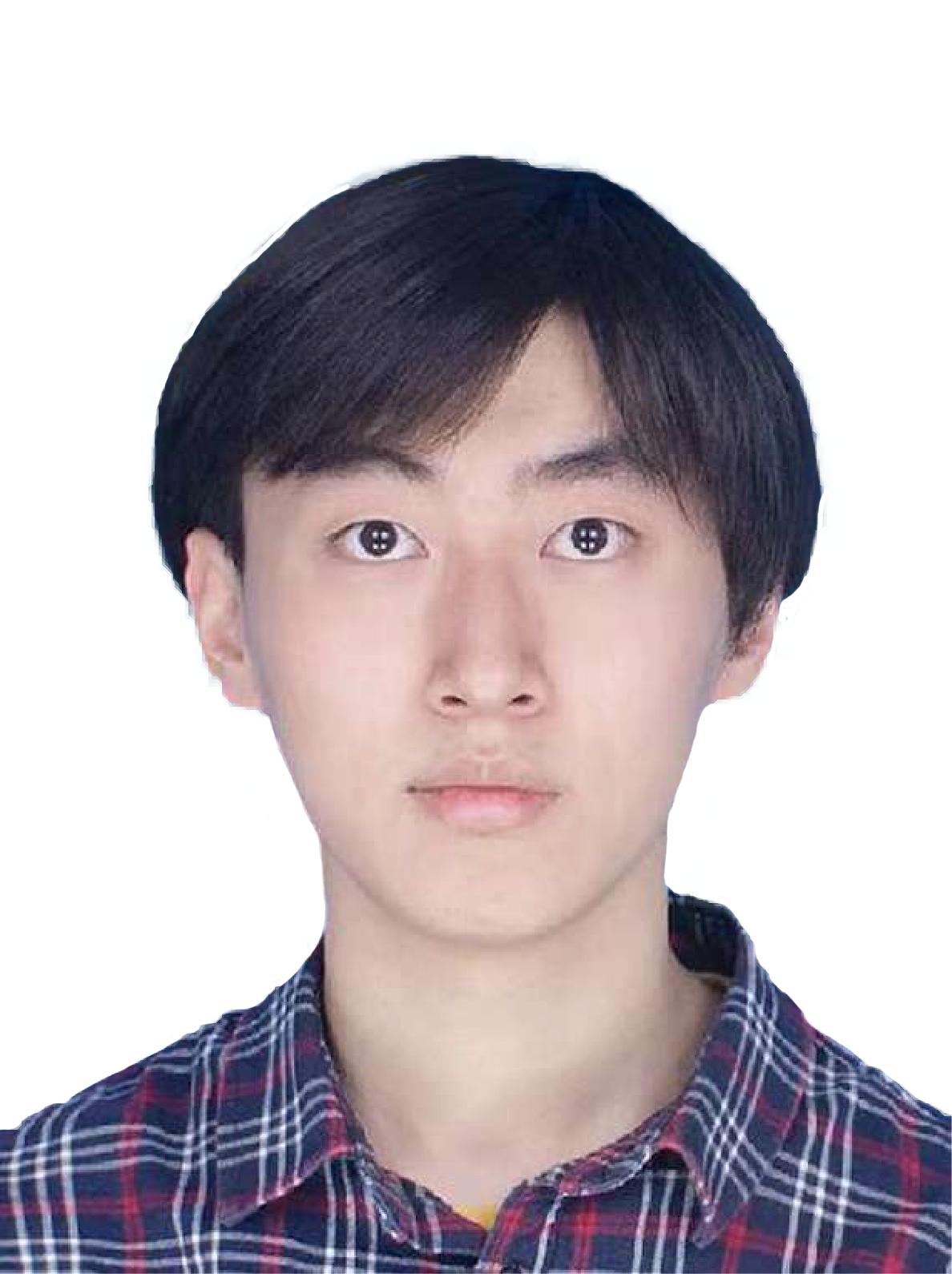}}]
{Liuzhuozheng Li} is currently pursuing the MS.C. degree in Complexity Science and Engineering at The University of Tokyo, Tokyo, Japan. His main research interests include the deep generative model and computer vision. \end{IEEEbiography}

\begin{IEEEbiography}[{\includegraphics[width=1in,height=1.25in,clip,keepaspectratio]{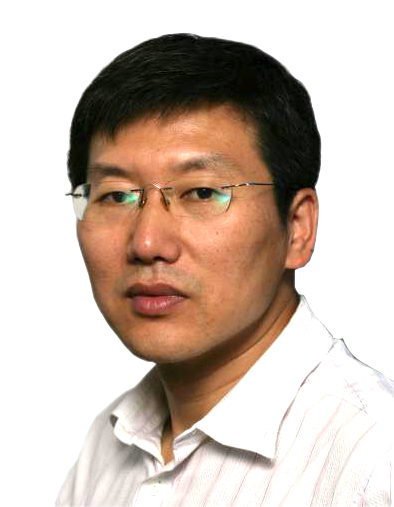}}]
{Hongyu Wang} (Member, IEEE) received his B.S. degree in electronic engineering from the Jilin University of Technology, Changchun, China, in 1990, the M.S. degree in electronic engineering from the Graduate School, Chinese Academy of Sciences, Beijing, China, in 1993, and the Ph.D. degree in precision instrument and optoelectronics engineering from Tianjin University, Tianjin, China, in 1997. He is currently a Professor with the Dalian University of Technology, Dalian, China. His research interests include image processing, computer vision, 3-D reconstruction, and simultaneous localization and mapping (SLAM). \end{IEEEbiography}

\begin{IEEEbiography}[{\includegraphics[width=1in,height=1.25in,clip,keepaspectratio]{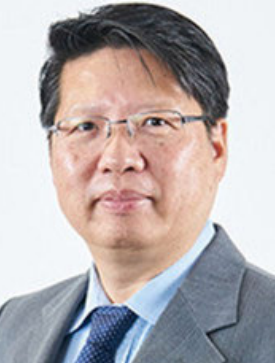}}]
{Danwei Wang} (Life Fellow, IEEE) received the B.E. degree from the South China University of
Technology, China, in 1982, and the M.S.E. and Ph.D. degrees from the University of Michigan, Ann Arbor, MI, USA, in 1984 and 1989, respectively. He is a fellow of the Academy of Engineering Singapore. He was a recipient of the Alexander von Humboldt Fellowship, Germany. He served as the general chairperson, the technical chairperson, and various positions for several international conferences. He was an invited guest editor of various international journals. He is a Distinguished Lecturer of the IEEE Robotics and Automation Society. \end{IEEEbiography}

\begin{IEEEbiography}[{\includegraphics[width=1in,height=1.25in,clip,keepaspectratio]{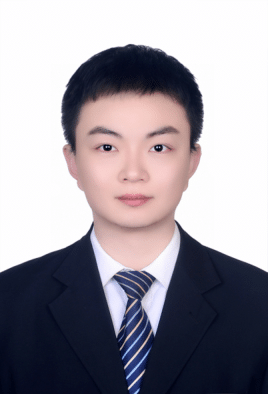}}]
{Mingrui Li} is currently pursuing the Ph.D. degree with the School of Information and Communication Engineering, Dalian University of Technology, Dalian, China. His research interests include 3D reconstruction, simultaneous localization and mapping (SLAM), and computer vision. \end{IEEEbiography}

\end{document}